\documentclass[12pt]{article}

\usepackage{mathrsfs}
\usepackage{natbib,graphicx,setspace,lscape,longtable}
\usepackage{mathrsfs,amsmath,amsthm,amssymb,color}
\usepackage{natbib,epsfig,graphicx,pdfpages}
\usepackage{rotating}
\usepackage{float}
\usepackage{bm}
\usepackage{ulem}
\usepackage{CJK}
\usepackage{natbib}
\usepackage{algorithm}
\usepackage{algorithmic}
\usepackage{booktabs}
\usepackage{bm}
\usepackage{subfigure}
\usepackage{makecell}
\usepackage{hyperref}
\usepackage{multirow}
\usepackage{threeparttable}
\usepackage{xr}

\hypersetup{
	colorlinks=true,
	linkcolor=blue,
	citecolor=blue,
	urlcolor=blue}

\bibpunct{(}{)}{;}{a}{,}{,}

\newtheorem{theorem}{Theorem}
\newtheorem{lemma}{Lemma}
\newtheorem{proposition}{Proposition}
\newtheorem{corollary}{Corollary}

\newcommand{\csection}[1]
    {\begin{center}
        \stepcounter{section}
        {\bf\large\arabic{section}. #1}
    \end{center}
}

\newcommand{\csubsection}[1]{
\begin{center}
\stepcounter{subsection}
{\it\arabic{section}.\arabic{subsection}. #1}
\end{center}
}

\def\n{\nonumber}

\def\beq{\begin{equation}}
\def\eeq{\end{equation}}
\def\beqr{\begin{eqnarray}}
\def\eeqr{\end{eqnarray}}
\def\beqrs{\begin{eqnarray*}}
\def\eeqrs{\end{eqnarray*}}
\def\bet{\begin{theorem}}
\def\eet{\end{theorem}}
\def\bel{\begin{lemma}}
\def\eel{\end{lemma}}
\def\bep{\begin{proposition}}
\def\eep{\end{proposition}}
\def\bg{\begin{figure}[tbph]\begin{center}}
\def\eg{\end{center}\end{figure}}

\def\bc{\begin{center}}
\def\ec{\end{center}}

\def\wh{\widehat}

\def\mR{\mathbb{R}}
\def\mS{\mathcal S}

\def\mL{\mathcal L}

\def\mlambda{\lambda_{\operatorname{max}}}
\def\1{\mbox{\boldmath $1$}}
\def\SEW{\widehat{\operatorname{SE}}(W)}
\def\SEL{\widehat{\operatorname{SE}}( \dot{\mL}(\theta_0)  )}

\def\SSEW{\widehat{ \operatorname{SE}}^2 (W)}

\def\SEX{\widehat{ \operatorname{SE}} (\widehat{\Sigma}_{xx})}
\def\SEY{\widehat{ \operatorname{SE}} (\widehat{\Sigma}_{xy})}
\def\thetaopt{\widehat{\theta}_{\operatorname{ge}}}
\def\thetaolss{\widehat{\theta}^*_{\operatorname{ols}}}

\def\mM{\mathcal M}

\def\argmin{\mbox{argmin}}

\numberwithin{equation}{section}

\begin{document}
\begin{center}
    {\bf\Large Network Gradient Descent Algorithm for Decentralized Federated Learning}\\
   Shuyuan Wu$^1$, Danyang Huang$^{2,3}$, and Hansheng Wang$^1$

   {\it $^1$Guanghua School of Management, Peking University, Beijing, China}\\
   {\it $^2$Center for Applied Statistics, Renmin University of China, Beijing, China}\\
   {\it $^3$School of Statistics, Renmin University of China, Beijing, China}\\
   \begin{footnotetext}[1]
{
Danyang Huang's research is partially supported National Natural Science Foundation of China (No. 12071477); fund for building world-class universities (disciplines) of Renmin University of China; Public Computing Cloud, Renmin University of China. Hansheng Wang's research is partially supported by National Natural Science Foundation of China (No. 11831008) and also  partially supported by the Open Research Fund of Key Laboratory of Advanced Theory and Application in Statistics and Data Science (KLATASDS-MOE-ECNU-KLATASDS2101).
}
\end{footnotetext}

    \end{center}
\begin{singlespace}
\begin{abstract}
 We study a fully decentralized federated learning algorithm, which is a novel gradient descent algorithm executed on a communication-based network. For convenience, we refer to it as a network gradient descent (NGD) method. In the NGD method, only statistics (e.g., parameter estimates) need to be communicated, minimizing the risk of privacy. Meanwhile, different clients communicate with each other directly according to a carefully designed network structure without a central master. This greatly enhances the reliability of the entire algorithm. Those nice properties inspire us to carefully study the NGD method both theoretically and numerically. Theoretically, we start with a classical linear regression model. We find that both the learning rate and the network structure play significant roles in determining the NGD estimator's statistical efficiency.  The resulting NGD estimator can be statistically as efficient as the global estimator, if the learning rate is sufficiently small and the network structure is well balanced, even if  the data are distributed heterogeneously. Those interesting findings are then extended to general models and loss functions. Extensive numerical studies are presented to corroborate our theoretical findings.  Classical deep learning models are also presented for illustration purpose.\\


\end{abstract}
\end{singlespace}

\newpage

\csection{INTRODUCTION}

We study here a gradient descent algorithm for decentralized federated learning. Our methodology has two important elements. The first is gradient descent, which is arguably the most popularly used optimization method for complex statistical learning (e.g. deep learning). The second is federated learning, which is a novel collective learning paradigm. The objective here is to train a global model collectively by a large number of local devices (or clients). Often those devices can be naturally connected with each other through a network (e.g. a wireless communication network). Then, how to conduct a gradient descent algorithm on this network and study the resulting estimator's statistical properties become problems of great interest.

Specifically, we consider a standard statistical learning problem with a total of $N$ observations. For each observation, there is a response of interest and a predictor
with a fixed dimension. There are also parameters of interest, which need to be estimated by minimizing an appropriately defined loss function. Traditionally,
the whole sample is placed on one single computer and then processed conveniently. However, in federated learning, data are often distributed across a large number of clients (e.g. mobile devices). One major concern is the privacy issue. Typically, the data generated from different clients could contain highly sensitive private information. Passing the original data from local clients to a central computer may incur a high risk of privacy disclosure. Nevertheless, to train the global model, we need the information contained in the datasets distributed across different clients. This inspires the novel idea of federated learning to fix the privacy disclosure problem.

The original idea of federated learning was first proposed by \cite{konevcny2016federated}. They assumed that there exists a central computer, which can be connected with a large number of clients. To train the global model, the central computer collects the local parameter estimators computed on each local client after local gradient updating iteratively. Then, the aggregating parameter estimators are passed to each local client.  Transporting local parameter estimators protects privacy better than directly passing the raw data from local clients to the central computer. This leads to the classical federated learning algorithm \citep{mcmahan2017communication}, which has been extended by many researchers. For example, \cite{smith2017federated} proposed a novel strategy to handle federated multi-task learning problems.
\cite{cheu2019distributed} proposed a distributed differentially private algorithm for stricter privacy protection. \cite{hashimoto2018fairness} developed a robust optimization method to handle worst-case risk.

These federated learning algorithms require a central machine, which is responsible for communicating with every client. Such  an  type of architecture is easy to implement. However, it suffers from several serious limitations. First, it is not the best choice for privacy protection, because the central machine itself could be vulnerable. In fact, if the central machine is conquered, the attacker is given the chance to communicate with every client \citep{bellet2018personalized}. Second, this centralized network structure is extremely fragile for stable operation. In other words, the central machine has an overly important role in the network. When the central machine stops working, the entire learning task stops. Third, centralized network structure has a high requirement for network bandwidth. This is because the central machine has to communicate with a huge number of clients \citep{li2020blockchain}.

To fix these problems, a number of researchers advocate the idea of fully decentralized federated learning \citep{colin2016gossip,vanhaesebrouck2017decentralized,tang2018d}. The key feature is that there is no central computer involved for model training and communication. Although one central computer may be needed to set up the entire learning task, it is not responsible for actual computation. Consequently, all the computation-related communications should occur only between individual clients. This leads to a communication-based network structure. Each node represents a client and each edge represents the communication relationship between the clients. Subsequently, local gradient steps are conducted for each client. The corresponding local parameters are updated by aggregating the information from their one-hop neighbours (i.e. directly connected neighbours) in the network.   This leads to the Gossip SGD method of \cite{blot2016gossip} and Decentralized (Stochastic-)GD method of  \cite{yuan2016convergence}, \cite{lian2017can}, and \cite{lian2018asynchronous}.
\cite{nedic2017achieving} and \cite{savazzi2020federated} further proposed  new algorithms based on  the exchange of both parameters and gradients in each iteration for  better convergence speed.
In addition,
\cite{lalitha2018fully,lalitha2019peer} proposed a Bayesian-type approach to estimate the interested parameter for  additive models in the aperiodic  strongly connected network.
\cite{richards2019optimal} and \cite{richards2020decentralised}
analyzed the statistical rates of Decentralized GD based on non-parametric regression with a squared loss function.

The above literature about decentralized federated learning could be summarized from different aspects according to the: (1)  assumptions imposed on the network structure; (2) assumptions imposed on the data distribution across different clients; and (3) types of  theoretical convergence; see Table \ref{tab:Explanation} for the details. By Table \ref{tab:Explanation}, we find that many existing literature imposed stringent assumptions on the network structure, e.g., doubly stochastic matrix \citep{yuan2016convergence,tang2018d,lian2017can,richards2019optimal,richards2020decentralised}. In addition, some literature imposed restrictive assumptions on the data distribution pattern, e.g. homogeneous distribution \citep{richards2019optimal,richards2020decentralised}. Moreover, most existing literature studied the algorithm convergence either theoretically or numerically. It seems that little has been done about the statistical convergence theory.

\begin{table}[h]\small
\renewcommand\arraystretch{1.3}
\setlength\tabcolsep{3pt}
\centering
\caption{\label{tab:Explanation}
The summarization of existing literature from three aspects according to the assumptions on the network structure, data distribution pattern, and types of theoretical convergence. The  definition of the abbreviations are given as follows. DSM:  doubly stochastic matrix;  SC: strongly connected;  s-SM: symmetric similarity matrix (i.e., the matrix value reflects distribution similarity);  WBM: weakly balanced matrix; BV: bounded variance (i.e., the variance of the stochastic gradient is bounded across different clients). HOMO: homogeneous (i.e., independent and identically distributed across different clients);  HETE: heterogeneous; SEB: statistical error bound; NCR/SCR: numerical/statistical convergence rate. Moreover,  None represents no stringent assumptions.
}
\begin{tabular}{cccc}
\hline
Existing & Network Structure  & Data Distribution  & Theoretical     \\
Literature& Assumption &  Assumption  & Convergence Type\\
\hline
\cite{yuan2016convergence}& \multirow{2}{*}{DSM} & \multirow{2}{*}{HOMO/HETE}& \multirow{2}{*}{NCR} \\
\cite{tang2018d}&\\
\hline
\cite{nedic2017achieving}& {DSM/SC} & HOMO/HETE&NCR\\
\cite{lian2017can,lian2018asynchronous}& {DSM} & {BV}& {NCR} \\
\cite{richards2019optimal}& \multirow{2}{*}{DSM} & \multirow{2}{*}{HOMO}& \multirow{2}{*}{NCR\&SCR} \\
\cite{richards2020decentralised}&\\
\hline
\cite{vanhaesebrouck2017decentralized}&
s-SM&HOMO/HETE&
NCR\\
\cite{lalitha2018fully,lalitha2019peer}&aperiodic-SC&HOMO/HETE  & SEB \\
\hline
\cite{blot2016gossip} &
\multicolumn{3}{c}{\multirow{2}*{Algorithm Driven}}\\
\cite{savazzi2020federated} &  \\
\hline
\textbf{The Proposed NGD Method}&\textbf{ WBM}&\textbf{HOMO/HETE} & \textbf{NCR\&SCR} \\
\hline
\end{tabular}
\end{table}

In our work, we develop a novel methodology for decentralized federated learning. The new methodology allows the data distribution pattern across different clients to be either homogeneous or heterogeneous. Moreover, the proposed method only requires the network structure to be weakly balanced. This is an assumption weaker than the double stochastic matrix assumption, which has been popularly used in the past literature \citep{yuan2016convergence,tang2018d,richards2019optimal,richards2020decentralised}. Meanwhile,  both the numerical convergence and statistical efficiency are studied. Specifically, our methodology is based on the classical structure of the decentralized federated learning \citep{yuan2016convergence,blot2016gossip}. For convenience, we call the method network gradient descent (NGD) method. First, we assume that different clients are connected with each other by an appropriately defined network. The corresponding network structure can be mathematically described by a network adjacency matrix. Second, we conduct a gradient descent algorithm on each local client. To calculate the local gradients, the target client needs to update the current estimator to be the next one by the method of gradient descent. The gradient is computed based on the current estimator, which is taken as the averaged estimators generated by the network neighbours. This leads to an interesting gradient descent algorithm, which can be executed on a fully connected network.

We theoretically prove that the NGD algorithm numerically converges to a limiting estimator under appropriate conditions for the learning rate.
The resulting estimator is referred to as the NGD estimator and its  asymptotic properties are investigated.  We show theoretically that the statistical efficiency of the NGD estimator is  jointly affected by three factors.  They are, respectively, the learning rate, the network structure, and the data distribution across different clients.  Our theory suggests that a statistically efficient estimator can be obtained even if the data are heterogeneously distributed across different clients, as long as the learning rate is sufficiently small and the network structure is sufficiently balanced. This seems to us a very interesting or even surprising theoretical finding \citep{lian2017can,lian2018asynchronous}.  However, if the data distribution across different clients are well homogeneous, the technical requirement for both the learning rate and network structure can be further relaxed.
Extensive numerical experiments are conducted to  demonstrate our theoretical findings.

The rest of the article is organized as follows. Section 2 develops the NGD algorithm and presents its numerical and asymptotic statistical properties. The numerical studies are presented in Section 3, including the simulation experiments and a real data example of image data analysis by deep learning. Finally, Section 4 concludes with a brief discussion. All technical details are relegated to the Appendix.

\csection{METHODOLOGY}
\noindent

\csubsection{Network Gradient Descent}
\noindent

We first introduce the model and notations.
Let $X_i \in \mathbb{R}^p, Y_i \in \mathbb{R}^1$ be the information collected from the $i$th subject with $1 \leq i \leq N.$ Here $Y_i $ is the response of interest and  $X_i$ is the associated predictor  with finite moments. For simplicity, we start with the following classical linear regression model
\beqrs
Y_i = X_i^\top \theta_0 + \varepsilon_{i},
\eeqrs
where $\varepsilon_{i}$ is the independently and identically distributed residual with mean $0$ and
variance $\sigma^2$. Here $\theta_0$ is the regression coefficient parameter to be estimated. To estimate $\theta_0,$ we need to optimize the following global least-squares loss function as $\mL_{N}(\theta) = N^{-1} \sum_{i=1}^N (Y_i - X_i^\top \theta)^2.$ Then, it leads to the corresponding ordinary least squares (OLS) estimator $\widehat{\theta}_{\text{ols}} = \operatorname{argmin}_{\theta} \mL_N(\theta) = \widehat{\Sigma}_{xx}^{-1} \widehat{\Sigma}_{xy}$, where $\widehat{\Sigma}_{xx} = N^{-1} \sum_{i} X_i X_i^\top $ and $\widehat{\Sigma}_{xy} = N^{-1}\sum_{i} X_i Y_i.$ It is remarkable that based on the least squares loss function, we are able to obtain theoretical results with deep insights. Those results are then extended to general loss functions in Subsection 2.5.

By the classical linear regression theory, we know that $\widehat{\theta}_{\text{ols}}$ is asymptotic normal with  $\sqrt{N} (\widehat{\theta}_{\text{ols}} - \theta_0) \to_d N(0, \sigma^2 \Sigma_{xx}^{-1})$ with $\Sigma_{xx} = E(X_iX_i^\top)$; see \cite{rao1973linear} and \cite{shao1999mathematical}.
 Consequently, as long as $\widehat{\Sigma}_{xx}^{-1}$ and $\widehat{\Sigma}_{xy}$ can be computed directly, the global OLS estimator can be easily obtained. However, such a straightforward method is often practically infeasible in federated learning. This is mainly because that data are distributed across different clients. To compute $\widehat{\Sigma}_{xx}^{-1}\widehat{\Sigma}_{xy}$ directly, one must aggregate all the local data from different clients to the central computer. As mentioned in the introduction, this leads to high risk of privacy disclosure and is practically undesirable. As a result, various federated learning algorithms are practically attractive.

Specifically, we consider a fully decentralized federated learning framework with the whole data distributed across $M$ clients, which are also nodes in the network. The clients are indexed by $\mM = \{1,2,\dots,M\}$. Those clients are connected by a communication network, whose adjacency matrix is given by $A = (a_{m_1m_2}) \in \mathbb{R}^{M \times M}$ for $1\leq m_1,m_2\leq M$. Here, $a_{m_1m_2} = 1$ if client $m_1$ can receive information from $m_2$, and $a_{m_1m_2} = 0$ otherwise. For completeness, we assume $a_{mm}=0$ for $1\leq m\leq M$. We define the weighting matrix $W = (w_{m_1m_2}) \in \mathbb{R}^{M \times M}$ with $w_{m_1m_2} = a_{m_1m_2}/d_{m_1}$, where $d_{m_1} = \sum_{m_2} a_{m_1m_2}$ is the in-degree for client (i.e. node) $m_1$  and we assume $d_{m_1} \geq 1$. Define the whole dataset $\mS_F = \{1,2,\dots ,N\} = \bigcup_{m=1}^M  \mS_{(m)}$, where $\mS_{(m)}$ is the index set of the sample distributed to the $m$th client. Then, the global loss function could be rewritten as
$$\mathcal{L}_{N}(\theta)=N^{-1} \sum_{i=1}^{N}\Big(Y_{i}-X_{i}^{\top} \theta\Big)^{2}= M^{-1}\sum_{m=1}^{M} \mathcal{L}_{(m)}(\theta),$$
where $\mathcal{L}_{(m)}(\theta)= n^{-1} \sum_{i \in \mathcal{S}_{(m)}}\left(Y_{i}-X_{i}^{\top} \theta\right)^{2} $ is the loss function evaluated for the $m$th client. For simplicity, we assume that $|\mS_{(m)}| = n=N/M$.

We then introduce the network gradient algorithm as follows. Specifically, we follow the idea of  \cite{yuan2016convergence}, \cite{colin2016gossip}, and \cite{lian2017can}, and execute the following NGD algorithm:
\beqrs
\label{eq:iterate_form}
\widehat{\theta}^{(t+1, m)}=\widetilde{\theta}^{(t, m)}-\alpha \dot{\mathcal{L}}_{(m)}\big(\widetilde{\theta}^{(t, m)}\big),
\eeqrs
where $\widetilde{\theta}^{(t, m)}$ is the neighbourhood averaged estimator obtained in the $t$th iteration for the $m$th client. Specifically, we have $\widetilde{\theta}^{(t, m)}=\sum_{k=1}^{M} w_{m k} \widehat{\theta}^{(t, k)}$. The optimization process is a decentralized federated learning process. Write $\widehat{\Sigma}_{x x}^{(m)}=\sum_{i \in \mathcal{S}_{(m)}} X_{i} X_{i}^{\top} / n$ and $\widehat{\Sigma}_{x y}^{(m)}=\sum_{i \in \mathcal{S}_{(m)}} X_{i} Y_{i} / n$. Then, we have,
\beqr
\label{eq:NGD}
\widehat{\theta}^{(t+1,m)}&=&\widetilde{\theta}^{(t,m)}-\alpha \dot{\mathcal{L}}\left(\widetilde{\theta}^{(t,m)}\right)
= \widetilde{\theta}^{(t,m)}-\alpha\left(\widehat{\Sigma}_{x x}^{(m)} \widetilde{\theta}^{(t,m)}-\widehat{\Sigma}_{x y}^{(m)}\right) \nonumber \\
&=&\left(I_p-\alpha \widehat{\Sigma}_{x x}^{(m)}\right) \widetilde{\theta}^{(t,m)}+\alpha \widehat{\Sigma}_{x y}^{(m)} \nonumber \\
&=&\left(I_p-\alpha \widehat{\Sigma}_{x x}^{(m)}\right)\Big(\sum_{k=1}^{M} w_{m k} \widehat{\theta}^{(t,k)}\Big)+\alpha \widehat{\Sigma}_{x y}^{(m)},
\eeqr
where $I_p\in\mR^{p\times p}$ is a $p$-dimensional identity matrix.

The estimator obtained by the NGD algorithm on each client could be reorganized into a vector form. Then, we define $\widehat{\theta}^{*(t)}= \big(\hat{\theta}^{(t,1) \top}, \cdots,$ $ \hat{\theta}^{(t,M) \top}\big)^{\top} \in \mathbb{R}^{ q}, \widehat{\Sigma}_{x y}^{*}=\big(\widehat{\Sigma}_{x y}^{(1) \top}, \cdots, \widehat{\Sigma}_{x y}^{(M) \top}\big)^{\top} \in \mathbb{R}^{q},$ and $\Delta^* = \operatorname{diag} \big(\Delta^{(1)},\Delta^{(2)},\dots,\Delta^{(M)}   \big) $ $\in \mR^{q\times q}$ with $\Delta^{(m)} = I_p - \alpha \widehat{\Sigma}^{(m)}_{xx}$  and $q = M p$. Then, (\ref{eq:NGD}) can be rewritten into a matrix form as
\beq
\label{eq:NSGD_mat}
\widehat{\theta}^{*(t+1)}=\Delta^{*}(W \otimes  I_p) \widehat{\theta}^{*(t)}+\alpha \widehat{\Sigma}_{x y}^{*},
\eeq
where $\otimes$ denotes the Kronecker product.
By equation (\ref{eq:NSGD_mat}), we find that the NGD estimation process can be represented by a linear dynamic system. This representation immediately leads to two interesting questions. First, does a stable solution exist for this linear dynamic system? Second, if there is a stable solution, does the aforementioned NGD algorithm converge to it? These problems are considered in the next subsection.

\csubsection{Numerical Convergence}
\noindent

To study the numerical properties of the NGD algorithm, we temporarily assume that a stable solution indeed exists. We then study its analytical expression. This leads to sufficient conditions about its existence. Before we discuss the stable solution, we first introduce the notations. Define
$\widehat{\theta}^*$ as the stable solution of equation ({\ref{eq:NSGD_mat}}). Then, we have
\beq
\label{eq:stable_mat}
\widehat{\theta}^{*}=\Delta^{*}(W \otimes I_p) \widehat{\theta}^{*}+\alpha \widehat{\Sigma}_{x y}^{*}.
\eeq
Denote $\widehat{\Omega} = I_p - \Delta^{*}(W \otimes I_p) \in \mathbb{R}^{q \times q}.$ Its analytical form is given by
\beqrs
\widehat{\Omega}  = \left(\begin{matrix}
    I_p - w_{11}\Delta^{(1)}, & - w_{12}\Delta^{(1)}, &  \dots , & - w_{1M}\Delta^{(1)} \\
    - w_{21}\Delta^{(2)}, & I_p - w_{22}\Delta^{(2)}, &  \dots , &  - w_{2M}\Delta^{(2)}\\
    - w_{31}\Delta^{(3)}, & - w_{32}\Delta^{(3)}, &  \dots , &  - w_{3M}\Delta^{(3)}\\
    \dots, & \dots, &  \dots , & \dots\\
    - w_{M1}\Delta^{(M)}, &  - w_{M2}\Delta^{(M)}, &  \dots , & I_p - w_{MM}\Delta^{(M)}\\
\end{matrix} \right) .
\eeqrs
If $\widehat{\Omega}$ is invertible, the stable solution is uniquely determined, and it is given by $\widehat{\theta}^* = \alpha \widehat{\Omega}^{-1} \widehat{\Sigma}^*_{xy}.$ Accordingly, the invertibility of $\widehat{\Omega}$ determines the existence of the stable solution $\widehat{\theta}^*$. If $\widehat{\theta}^*$ can indeed be obtained by the NGD algorithm ({\ref{eq:NSGD_mat}}), we refer to it as the NGD estimator.

Next, we discuss the convergence conditions of the NGD algorithm. To this end, we take the difference between equations (\ref{eq:NSGD_mat}) and (\ref{eq:stable_mat}). This leads to the following equation
\beq
\label{eq:diff}
\widehat{\theta}^{*(t+1)} - \widehat{\theta}^{*} = \Delta^{*}(W \otimes I_p) \Big( \widehat{\theta}^{*(t)} -  \widehat{\theta}^{*} \Big).
\eeq
We refer to $\Delta^{*}(W \otimes I_p)$ as a contraction operator. As (\ref{eq:diff}) shows, the contraction operator plays an important role in determining the numerical
convergence of the NGD algorithm. Specifically, we should have $\widehat{\theta}^{*(t+1)} - \widehat{\theta}^{*}  = \big\{ \Delta^{*}(W \otimes I_p)  \big\}^{t+1} ( \widehat{\theta}^{*(0)} -  \widehat{\theta}^{*})$ by (\ref{eq:diff}), where $\widehat{\theta}^{*(0)}$ is the initial value specified for every client. Then, for the numerical convergence of the NGD algorithm, we should require
$\mlambda\big(\Delta^{*}\{W \otimes I_p\}\big) < 1$, where $\lambda_{\max}(B)$ is the largest absolute eigenvalue of an arbitrary matrix $B$. Notice that the eigenvalue of $B$ could be a complex number, if $B$ is asymmetric. Moreover, under the condition $\mlambda\big(\Delta^{*}\{W \otimes I_p\}\big)  < 1,$ $\widehat\Omega$ is also invertible. Thus, the stable solution $\widehat{\theta}^*$ (now equal to the NGD estimator) does exist. Finally, define a sequence $\{x^t\}$ as converging linearly if there are some $t_0 \in \mathbb{N}$ such that $\|x^t\| \leq c_0 \nu^t$ for all $t > t_0$, with  some constants $\nu \in [0,1)$ and $c_0 > 0.$
We then obtain the following theorem to assure the numerical linear convergence of the NGD algorithm.

\begin{theorem}
    \label{prop:stable_condi}
   (Numerical Convergence for Linear Regression) Assume that $n>p$ and $0 < \alpha < 2\min_{1 \leq m \leq M} \big\{\mlambda^{-1}(\widehat{\Sigma}_{xx}^{(m)}) \big\}$. Then, the stable solution $\widehat{\theta}^*$ exists, and  $\widehat{\theta}^{*(t)} \to \widehat{\theta}^{*}$ linearly.
\end{theorem}

\noindent
The detailed proof of Theorem \ref{prop:stable_condi} is given in Appendix B.1. Based on Theorem \ref{prop:stable_condi}, we draw the following two important conclusions. First, as long as the learning rate $\alpha$ is sufficiently small, and the sample size $n$ of each client is larger than the dimension of parameter $p$, the stable solution $\widehat{\theta}^*$ should exist, and the NGD algorithm should converge to it linearly. This leads to the NGD estimator. Nevertheless, we should note that there exists a situation, where $\widehat\Omega$ is invertible but $\mlambda\big(\Delta^{*}\{W \otimes I_p\}\big)  > 1.$ In this case, the stable solution exists, but the NGD algorithm might not converge to it. Thus, the NGD estimator does not exist because it cannot be computed. Second, surprisingly, we find that the numerical convergence of the NGD algorithm can be fully assured by the learning rate only. In other words, the network structure is not very important in determining the NGD algorithm's numerical convergence. However, as we show subsequently, the network structure does play a critical role in determining the NGD estimators' statistical efficiency. This is demonstrated in the next subsection.

\csubsection{The NGD Estimator versus the OLS Estimator}
\noindent

By the results given in Subsection 2.2, we know that, under appropriate regularity conditions, the NGD algorithm should numerically converge to the NGD estimator $\widehat\theta^*$. Then, it is of great interest to investigate its statistical efficiency. To this end, the relationship between $\widehat{\theta}^*$ and the global estimator (i.e. the OLS estimator $\widehat{\theta}_{\operatorname{ols}}$), is investigated. Write $\widehat{\theta}^*_{\operatorname{ols}} = I^* \widehat{\theta}_{\operatorname{ols}}$ as the stacked global OLS estimator, where $I^* = \1_{M} \otimes I_p$, and $\textbf{1}_M=(1,\cdots,1)^\top \in\mR^M$. Furthermore, define $\widehat{ \operatorname{SE}}^2 (W) = M^{-1}\| W^\top \textbf{1}_M - \textbf{1}_M \|^2$, where $\|v\| = (v^\top v)^{1/2}$ for an arbitrary vector $v.$ Note that the mean for the column sum of $W$ is 1. Thus, $\widehat{ \operatorname{SE}}^2 (W) $ measures the variability of the column sum of $W$. Intuitively, $\widehat{ \operatorname{SE}}^2 (W) $ can be considered as a measure for the balance of the network structure. A small $\widehat{ \operatorname{SE}}^2 (W)$ value suggests that different clients are connected with each other with approximately equal likelihood. For example, if the network structure is of a doubly stochastic structure \citep{yuan2016convergence,tang2018d}, then $ \widehat{ \operatorname{SE}}^2 (W)=0.$ This implies a perfectly balanced network structure. Otherwise, some clients must be overly preferred by the network, which means imbalance to some extend. In this regard, several classical network structures should be discussed in the following Subsection 2.4. Write
$\widehat{ \operatorname{SE}}^2 (\widehat{\Sigma}_{xx}) = \operatorname{tr} \Big[ M^{-1}\sum_{m=1}^M ( \widehat{\Sigma}^{(m)}_{xx} - \widehat{\Sigma}_{xx} ) ^2 \Big]$ and
$\widehat{ \operatorname{SE}}^2 (\widehat{\Sigma}_{xy}) = M^{-1}\sum_{m=1}^M \| \widehat{\Sigma}^{(m)}_{xy} - \widehat{\Sigma}_{xy} \|^2$.  If observations are distributed across different clients in a completely random way, we expect $ \widehat{ \operatorname{SE}}^2 (\widehat{\Sigma}_{xx})$ and $ \widehat{ \operatorname{SE}}^2 (\widehat{\Sigma}_{xy})$ to be of  $O_p(1/n)$ order;  Otherwise, the data distribution across different clients should be heterogeneous; see Lemma 3 in Appendix A for the detailed proof.  Thus, both $ \widehat{ \operatorname{SE}}^2 (\widehat{\Sigma}_{xx})$ and $ \widehat{ \operatorname{SE}}^2 (\widehat{\Sigma}_{xy})$ can be viewed as measures for data distribution randomness.  Finally, define $\lambda_{\min}(B)$, $\lambda^+_{\min}(B)$ as the smallest eigenvalue and the smallest positive eigenvalue of an arbitrary matrix $B,$ respectively. Let  $ \sigma^w_{\max} = \lambda_{\max}^{1/2} (W^\top W),$ $\sigma_{\min}^{I-w} = (\lambda_{\min}^{+ } \big\{(I_p -W)^\top (I_p -W) \big\})^{1/2}.$  We then obtain the following theorem.

\begin{theorem}
\label{thm:numerical_converge}
 (Statistical Efficiency for Linear Regression) Assume the stable solution $\widehat{\theta}^*$ exists. Further, assume that (1) $W$ is irreducible, (2) there exist some positive constants $0 < \kappa_1 \leq \kappa_2 < \infty,$ such that   $ \kappa_1 \leq \lambda_{\min}\big( \widehat{\Sigma}_{xx} \big)\leq  \mlambda\big( \widehat{\Sigma}_{xx}^{(m)} \big) \leq \kappa_2$ for any $1 \leq m \leq M$,
(3) $\SEW $ and $\alpha $ are sufficiently small such that $\alpha \kappa_2 \sigma_{\max}^w + \SEW <  \kappa_1 \kappa_2^{-1} \sigma_{\min}^{I-w} /(4\sigma_{\max}^w). $ Then it holds that
\beqrs
 \frac{\|\widehat{\theta}^* -  \thetaolss \|}{\sqrt{M}} &\leq& c_1 \big\{\SEW + \alpha \big\} \Big[ \SEX   + \SEY  \Big]
\eeqrs
for some constants $c_1 > 0$ with probability tending to one.
\end{theorem}
\noindent

\noindent
The detailed proof of Theorem \ref{thm:numerical_converge} is given in Appendix B.2.
By Theorem \ref{thm:numerical_converge}, the discrepancy between $\wh{\theta}^*$ and $\wh{\theta}^*_{\operatorname{ols}}$ is upper bounded by $\big\{ \SEW + \alpha\big\}\big\{\SEX + \SEY\big\} $. As a result, the optimal statistical efficiency can be guaranteed, as long as this term is of  $o_p(1/\sqrt{N})$ order. This can be easily satisfied by: either (1) $\alpha=o_p(1/\sqrt{N})$ and $\SEW=o_p(1/\sqrt{N})$; or (2) $\SEX=O_p(1/\sqrt{n})$, $\SEY=O_p(1/\sqrt{n})$ and   $\alpha=o_p(1/\sqrt{M})$, $\SEW=o_p(1/\sqrt{M})$. Note that (2) represents the case where data cross different clients are homogeneously, which means identically and independently distributed.
In this case, $M$ can play its role through $\SEX+\SEY = O_p(\sqrt{M/N})$. This suggests that a larger $M$ leads  to a larger value for $\SEX+\SEY$,  and then a worse statistical convergence rate.  It is remarkable that a doubly stochastic matrix has been  typically assumed in previous literature \citep{yuan2016convergence,tang2018d,richards2019optimal,richards2020decentralised}. It requires $\SEW=0$.  However, here we only require that $\SEW$ to be relatively small, i.e., $o_p(1/\sqrt{M})$ or $o_p(1\sqrt{N})$. For convenience, we define $W$ that satisfies this condition to be a {\it weakly balanced network structure}.

To summarize, by Theorem \ref{thm:numerical_converge}, the statistical efficiency of the final estimator is determined by the following three factors: (1) the learning rate,  (2) the network structure, and (3) the data distribution pattern. Regarding learning rate, we find that the statistical efficiency of $\wh {\theta}^*$ improves as $\alpha$ decreases. Regarding $W$, we find that $\wh {\theta}^*$ becomes more efficient if $W$ is more balanced (e.g. $W$ is of a doubly stochastic structure). Regarding data distribution pattern, we find that homogeneous data distribution  leads to smaller distance between the  resulting estimator and $\thetaolss$.

When $\alpha$ is fixed or $W$ is arbitrarily specified, the NGD algorithm might not converge numerically to the global OLS estimator at all, even if the data are distributed independently and identically among different clients. This is particularly true if the sample size of each client is relatively small and the number of clients is relatively large \citep{kairouz2019advances}. Consequently, to obtain an NGD estimator as efficient as the OLS estimator,  both $\alpha$ and $\SEW$ are critically important. To this end, the learning rate $\alpha$ should be sufficiently small and the network structure should be well balanced.  This inspires us to study different network structures in this regard.

\csubsection{Network Structures}

As demonstrated in the previous subsection, the network structure plays a very important role in the statistical efficiency of the proposed NGD method through $\SEW$.
In this subsection, we study the $\SEW$ values of a number of important network structures. Specifically, we consider here three important network structures: the central-client network, the circle-type network, and the fixed-degree network.

\begin{figure}[h]
    \centering
	\setlength{\abovecaptionskip}{1pt}
	\subfigure[Central-Client]{\includegraphics[width=1.9in]{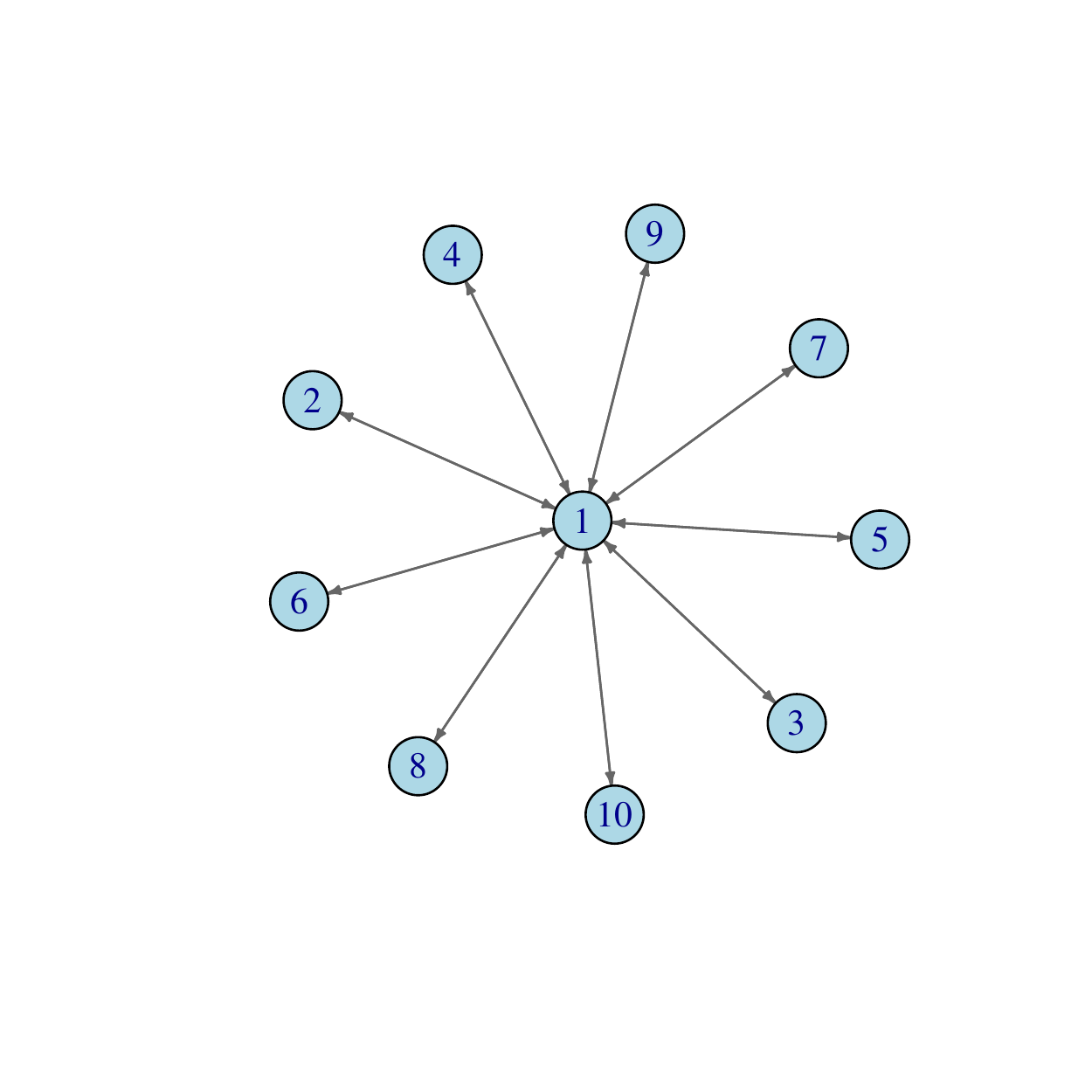}}
    \subfigure[Circle-Type]{
        \includegraphics[width=1.9in]{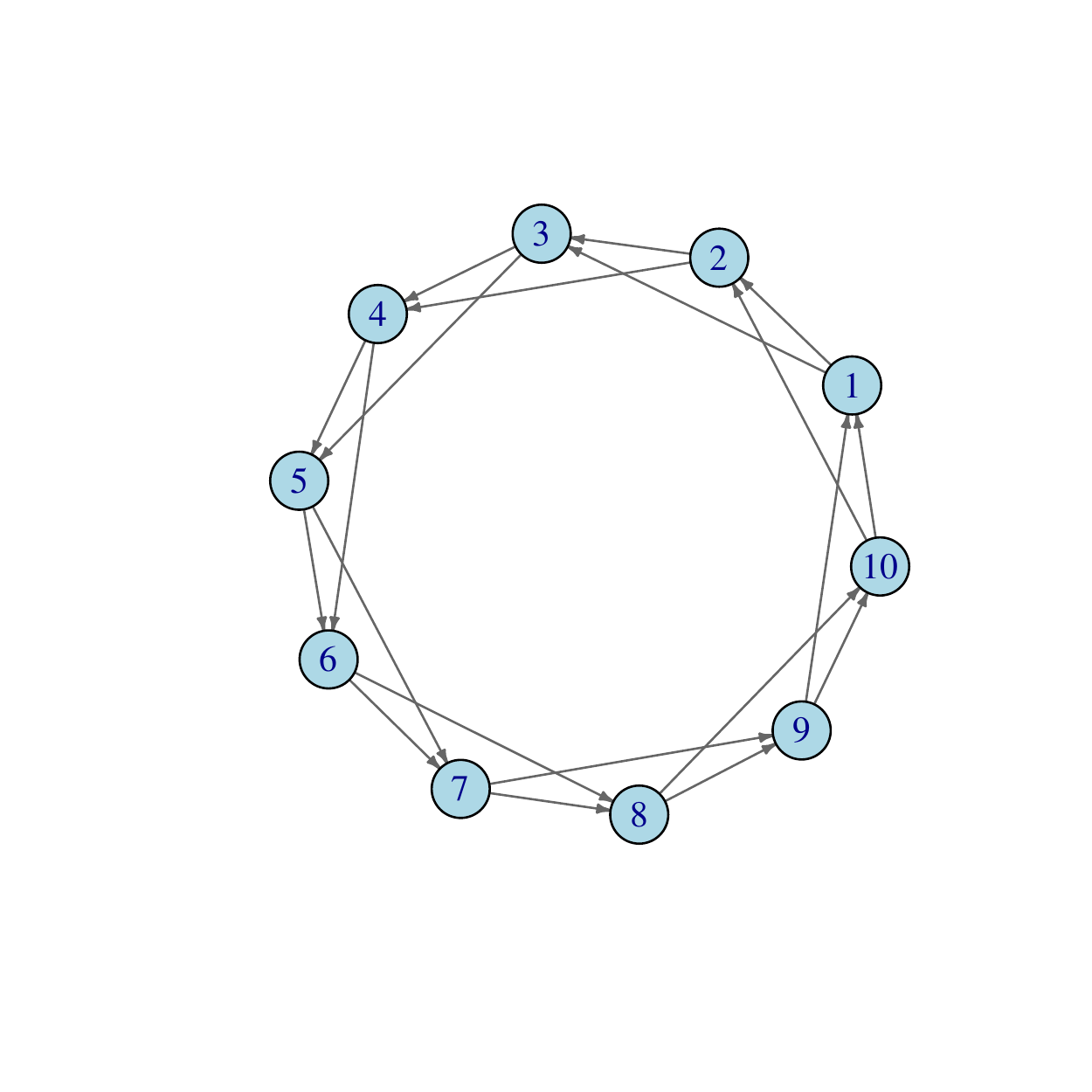}}
    \subfigure[Fixed-Degree]{
        \includegraphics[width=1.9in]{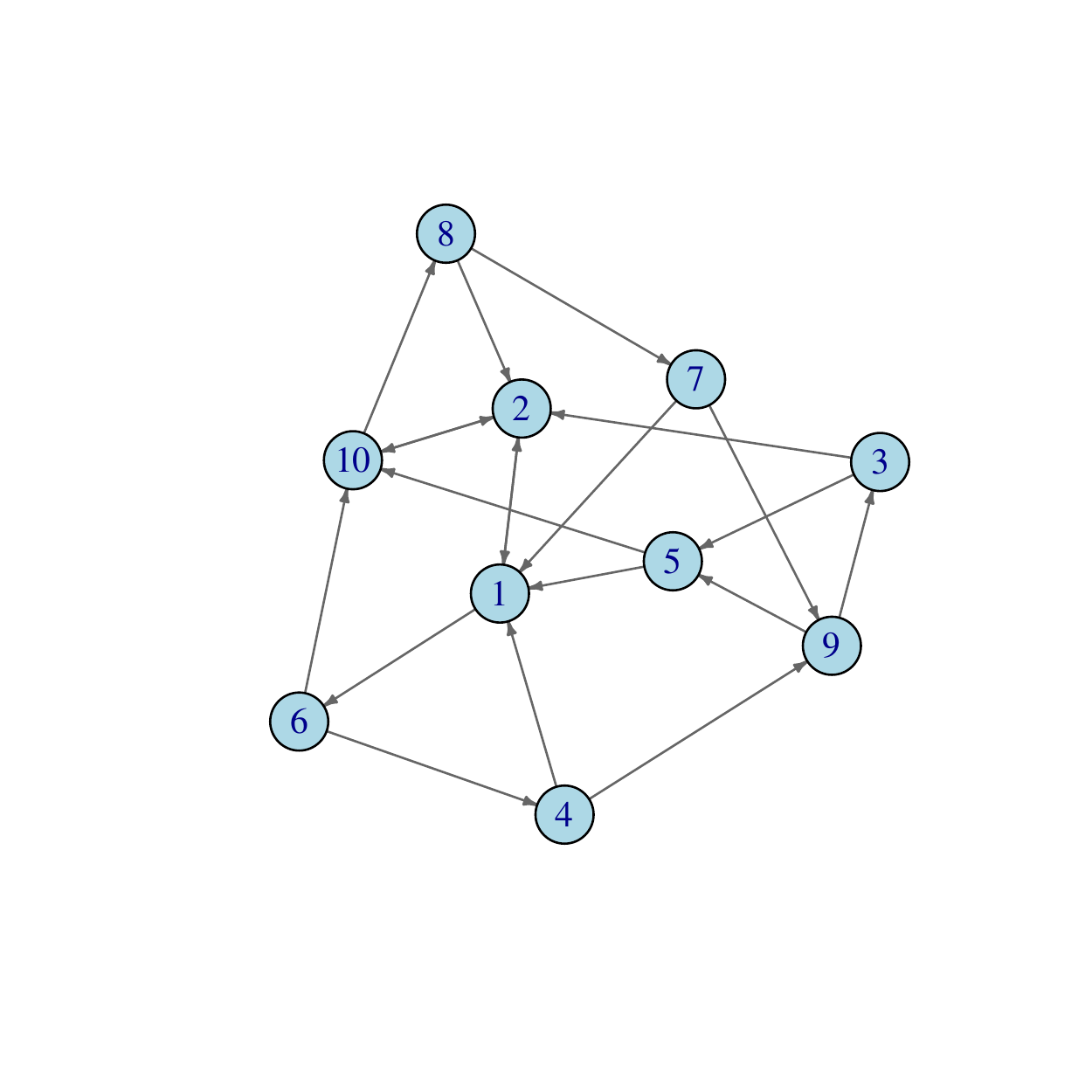}}
    \caption{Examples of three important network structures. Left Panel: the central-client network; middle panel: the circle-type network; and right panel: the fixed-degree network.}
    \label{fig:1}
\end{figure}

{\sc Case 1. (Central-Client Network)} We first consider the most typically used central-client structure. Without loss of generality, we assume that the first client with $m_1=1$ is the central computer, which should be
connected with all the other clients. However, not all the other clients are connected with each other. This leads to the network structure $A=(a_{m_1 m_2})\in\mR^{M\times M}$ with $a_{1 m_2}=a_{m_1 1}=1$ for every  $m_1,m_2\neq 1$ and $a_{m_1 m_2}=0$ otherwise.
Accordingly, we have the weighting matrix $W= (w_{m_1 m_2})\in\mR^{M\times M}$ given by
$W  = [0,(M-1)^{-1} \1^\top_{{_{M-1}}};\1_{{_{M-1}}},\textbf{0}].$ See the left panel of Figure \ref{fig:1} for an example with $M=10$.
Next, we evaluate $\widehat{\operatorname{SE}}(W)$. It can be verified that  $\SSEW = (M-1)^{-1} (M-2)^2.$ The details of this derivation are given in Appendix B.3, showing that the resulting estimator should be inconsistent when $M > 2,$ and its performance could be even worse as $M$ increases.

{\sc Case 2. (Circle-Type Network)} In this case, assume the clients are already appropriately ordered with a fixed in-degree $d_{m_1} = D>0$ for any $1 \leq m_1 \leq M.$ Specifically, define a network structure
as $A=(a_{m_1m_2})\in\mR^{M\times M}$ with $a_{m_1m_2}=1$ if $m_2= \big\{(m_1+d-1)  $ mod $M\big\} +1 $ for $1 \leq d \leq D$, where $a$ mod $b$ represents the remainder after dividing $a$ by $b.$ Otherwise, we define $a_{m_1m_2}=0$. The resulting network structure should be
of a circle type. See the middle panel of Figure \ref{fig:1} for an example with $M=10$ and $D=2$. The associated $W$ matrix is given by
\beqr
\label{eq:circle}
W  = \left(\begin{matrix}
    0, & 1/D, &  1/D, & \dots ,& 0 \\
    0, & 0, &  1/D , & \dots ,& 0\\
    \dots, & \dots, &  \dots , &\dots ,& \dots\\
    1/D, &  1/D, &  1/D , &\dots ,& 0\\
\end{matrix} \right) .
\eeqr
Then, one can find $W$ satisfying $\sum_{i=1}^M w_{ij} = \sum_{j=1}^M w_{ij} = 1.$ This further implies that $\SSEW = 0.$ As demonstrated in Subsection 2.3, this is an extremely well-balanced network structure.
Nevertheless, it is challenging in practice to develop and maintain such a circle-type network structure,  when $M$ is relatively large.

{\sc Case 3. (Fixed-Degree Network)} We next consider a fixed degree but a random-sampled network structure. Specifically, we fix the in-degree $d_{m_1} = D>0$ for any $1 \leq m_1 \leq M$ with some pre-specified $D$. Next,  for each client $m$, we conduct simple random sampling without replacement by other clients. Write this sample as $\mM_{_D}$. We then define $a_{m_1m_2}=1$ if $m_2\in\mM_{_{D}}$ and 0 otherwise. This
leads to a network structure $A$,  which is then fixed throughout the whole iteration process. See the right panel of Figure \ref{fig:1} for an example with $M$=10.
Next, we investigate its $\SEW$ value. In this case, $\SEW$ is a random variable. Take the expectation on $\sum_{i=1}^M w_{ij},$ we have $E(\sum_{i=1}^M w_{ij}) = \sum_{i=1}^M E(w_{ij}) = \sum_{i=1, i \neq j}^M D^{-1} P(w_{ij}=1/D) = 1.$ We next compute the variance of $\sum_{i=1}^M w_{ij}$. Because $P(w_{ij} = D^{-1}) = P(w_{ij}^2 = D^{-2}) = D/(M-1)$ for any $i \neq j$  and $P(w_{ii} = 0) = P(w_{ii}w_{ij} = 0) = 1.$  Then, we have
$
\operatorname{SE}\big( \sum_{i=1}^M w_{ij} \big) = E\big\{\SSEW\big\} = E  \Big( \sum_{i=1}^M w_{ij}  - 1\Big)^2  = E  \Big( \sum_{i=1}^M w_{ij}  \Big)^2 - 1.
$
On the other hand, it can be verified (see Appendix B.3  for the details) that
\beqrs
E  \left( \sum_{i=1}^M w_{ij}  \right)^2 &=& \sum_{i=1}^M E\big(w_{ij}^2 \big) + \sum_{i \neq k}E\big(w_{ij} w_{kj} \big) \nonumber \\
&=& (M-1) \frac{1}{D^2} \frac{D}{M-1} + (M-1)(M-2)  \frac{1}{D^2} \frac{D^2}{(M-1)^2}\n\\
& =& \frac{1}{D} - \frac{1}{M-1} + 1. \nonumber
\eeqrs
Consequently, $\operatorname{SE}\big( \sum_{i=1}^M w_{ij} \big) = E\big\{\SSEW\big\}= D^{-1} - (M-1)^{-1} \ge 0 .$
The statistical efficiency of the fixed degree network may be slightly worse when $M$ is small than that in the circle-type network. However, fixed degree network is easy to be implemented in practice and is more robust against network attacks \citep{kairouz2019advances}.

\csubsection{General Loss Functions}

The abovementioned theoretical results are developed for the linear regression models and the OLS loss functions. We next extend those nice results to more general models and loss functions. In this regard, we assume $\mL_N(\theta)$ is a general loss function.  The true parameter $\theta_0 = \argmin_{\theta} E \big\{ \mL_N(\theta) \big\} $ and the global estimator $\thetaopt = \operatorname{argmin}_\theta \mL_N(\theta)$ are defined accordingly. Here we assume $\mL_N(\cdot)$, and $E\big\{\mL_N(\cdot)\big\}$ satisfy the standard regularity conditions in classical statistical analysis of $M$-estimators such that $\thetaopt - \theta_0 = O_p(1/\sqrt{N})$ \citep{van2000asymptotic,jordan2019communication}.
For example, $\mL_N(\theta)$ can be defined as a twice negative log likelihood function of a generalized linear model. Accordingly, $\thetaopt$ should be the maximum likelihood estimation (MLE). Then, the NGD algorithm can be executed as
\beqrs
\widehat{\theta}^{(t+1,m)}&=&\widetilde{\theta}^{(t,m)} -\alpha \dot{{\mathcal{L}}}_{(m)} \left(\widetilde{\theta}^{(t,m)}\right) \\
&=& \widetilde{\theta}^{(t,m)}-\alpha\bigg\{ \dot{\mL}_{(m)}(\thetaopt)  +  \ddot{\mL}_{(m)}(\widetilde{\xi}^{(t,m)})\big(\widetilde{\theta}^{(t,m)}-\thetaopt\big) \bigg\} \\
&=&\left\{I_p-\alpha \ddot{\mL}_{(m)}\big(\widetilde{\xi}^{(t,m)}\big)\right\} \Big(\sum_{k=1}^{M} w_{m k} \widehat{\theta}^{(t,k)}\Big)+\alpha \Big\{ \ddot{\mL}_{(m)}\big(\widetilde{\xi}^{(t,m)}\big)  \thetaopt - \dot{\mL}_{(m)}(\thetaopt)  \Big\} \nonumber ,
\eeqrs
where $\dot{\mL}_{(m)}(\theta)$ and $\ddot{\mL}_{(m)}(\theta)$ represents the  first- and second-order derivatives of $\mL_{(m)}(\theta)$ with respect to $\theta$.
The algorithm can be rewritten in a matrix form as
\beqrs
\widehat{\theta}^{*(t+1)}=\widetilde\Delta^{*(t)}\big(W \otimes  I_p\big) \widehat{\theta}^{*(t)}+\alpha\Big\{ \ddot{\mL}^* \big(\widetilde{\xi}^{(t)}\big)  \thetaopt^* - \dot{\mL}^*(\thetaopt)  \Big\} ,
\eeqrs
where   $\ddot{\mL}^*(\widetilde{\xi}^{(t)})= \operatorname{diag} \big(\ddot{\mL}_{(1)}(\widetilde{\xi}^{(t,1)}),\ddot{\mL}_{(2)}(\widetilde{\xi}^{(t,2)}),  \dots,$    $\ddot{\mL}_{(M)}(\widetilde{\xi}^{(t,M)})  \big)\in \mR^{q\times q},$ $\widetilde\Delta^{*(t)} = I_q - \alpha \ddot{\mL}^*(\widetilde{\xi}^{(t)}),$ $\dot\mL^*(\thetaopt) = \big( \dot\mL^\top_{(1)}(\thetaopt) ,\dot\mL^\top_{(2)}(\thetaopt) ,\dots,\dot\mL^\top_{(M)}(\thetaopt)  \big)^\top \in \mathbb{R}^q$ and $\thetaopt^* = I^* \thetaopt.$ Then, take the difference between $\widehat{\theta}^{*(t+1)}$ and the  global estimator $\thetaopt^*$, yielding
\beqr
\label{eq:glseq}
\widehat{\theta}^{*(t+1)} - \thetaopt^* =\widetilde\Delta^{*(t)}\big(W \otimes  I_p\big) \big\{ \widehat{\theta}^{*(t)} - \thetaopt^* \big\}- \alpha  \dot{\mL}^*(\thetaopt)   ,
\eeqr
Compare (\ref{eq:glseq}) for a general loss function with (\ref{eq:diff}), and obtain the following key differences. First, the contraction operator $\widetilde\Delta^{*(t)}\big(W \otimes  I_p\big)$ for a general loss changes as $t$ changes. Second, there is an extra remainder term in ({\ref{eq:glseq}}) due to $\dot{\mL}_{(m)}(\thetaopt) \neq 0$. Nevertheless, we should have $\sum_{m=1}^M \dot{\mL}_{(m)}(\thetaopt) = 0.$ All these differences make the corresponding theoretical investigation extremely challenging. Finally, denote $\Sigma_w = W^\top (I_M - M^{-1} \textbf{1}_M \textbf{1}_M^\top ) W,$  and write $\SEL = \left\{ M^{-1}  \sum_{m=1}^M \| \dot{\mL}_{(m)}(\theta_0) \|^2 \right\}^{1/2}$. Similarly with $\SEX$ and $\SEY$, we could treat $\SEL$ as a measure for data distribution randomness. If observations are independent and identically distributed across different clients, $\SEL$ should be of $O_p(1/n)$ order.  Otherwise, the value of $\SEL$ could be relatively  larger. Based on the above notations, we obtain the following theorem to describe the discrepancy between the resulting NGD estimator and the global estimator.
\begin{theorem}
\label{thm:gls}
 (Statistical Efficiency for General Loss Functions) Assume (C1) there are some positive constants $0 < \kappa_3 \leq \kappa_4 < \infty,$ such that $\kappa_3 \leq \lambda_{\min} (\ddot{\mL}_{(m)}(\theta)) \leq \lambda_{\max} (\ddot{\mL}_{(m)}(\theta)) \leq \kappa_4 $ for any $\theta \in \mathbb{R}^p$ and $1 \leq m \leq M;$   (C2) $\SEW $ and $\alpha $ are sufficiently small such that
$\mlambda^{1/2}(\Sigma_w) \leq \rho < 1$ and $\alpha \sigma_{max}^w \kappa_4 +\SEW < 0.5 (\kappa_4)^{-1} \kappa_3(1-\rho)$.
Then,  we have
\beq
\label{eq:gls}
\lim\limits_{t \to \infty} \big\|\widehat{\theta}^{*(t)} - \thetaopt^* \big\| /\sqrt{M} \leq  c_2 \Big\{ \widehat{ \operatorname{SE}} (W) +   \alpha \Big\} \SEL
\eeq
for some positive constants $c_2 > 0$ with probability tending to one.
\end{theorem}
\noindent
The detailed proof of Theorem {\ref{thm:gls}} is given in Appendix B.4. This theorem suggests that the good theoretical results for linear regression models and the OLS loss functions can be extended to more general models and loss functions. We find again that the distance between the NGD estimator $\widehat{\theta}^*$ and the global estimator $\thetaopt^*$ is linearly bounded by $\big\{\alpha + \SEW\big\}\SEL.$ To summarize, we find that the three factors (i.e., the learning rate, the network structure, and the data distribution) together as a whole  affect the convergence rate of the NGD estimator.  Specifically, a statistical efficient estimator can be obtained by either (1) $\alpha=o_p(1/\sqrt{N})$ and $\SEW=o_p(1/\sqrt{N})$; or (2) $\SEL=O_p(1/\sqrt{n})$ and   $\alpha=o_p(1/\sqrt{M})$, $\SEW=o_p(1/\sqrt{M})$.  

Theorem  \ref{thm:gls} studies the NGD algorithm under the global strong convexity and smoothness conditions. To further extend its applicability, we follow \cite{nesterov1998introductory} and \cite{jordan2019communication} and relax those technical conditions from the globally strong convexity to locally strong convexity. This leads to the following Corollary \ref{col:gls}. The detailed proof of Corollary {\ref{col:gls}} is given in Appendix B.5. We find that the key results of Theorem  \ref{thm:gls} remains valid.
\begin{corollary}
\label{col:gls}
 (Statistical Efficiency for Locally Strong Convex Functions) Assume  conditions  \noindent (C$1^\prime$): there exist some positive constants $0 < \kappa_3' \leq \kappa_4' < \infty,$ such that $\kappa_3' \leq \lambda_{\min} (\ddot{\mL}_{(m)}(\thetaopt)) \leq \lambda_{\max} (\ddot{\mL}_{(m)}(\thetaopt)) \leq \kappa_4',$
and $\|\ddot{\mL}_{(m)}(\theta) - \ddot{\mL}_{(m)}(\theta^\prime)\| \leq \kappa_4' \|\theta - \theta^\prime \| $ for all $\theta,\theta^\prime$ in the neighborhood of $\thetaopt$ and (C2) hold. Further assume that the initial value $\wh \theta^{(0,m)}$ lies close to $\thetaopt$, and $\alpha + \SEW$ is sufficiently small.
Then (\ref{eq:gls}) in Theorem \ref{thm:gls} holds.
\end{corollary}
\noindent
\textbf{Remark.} It is remarkable that Corollary {\ref{col:gls}} assumed locally strong convexity.
To deal with more general locally convex settings, a novel method has been developed by \cite{ho2020instability} and \cite{ren2022towards} for studying the estimation error and the computational complexity of algorithm-based estimators. The key idea is to construct  an interesting reference estimator sequence for the actual algorithm-based estimator sequence. The reference sequence is constructed based on the population version of the interested algorithm. Next, by studying this population version reference sequence, the optimization error of the interested algorithm can be investigated. Furthermore, by studying the difference between the actual and reference estimator sequences, the stability error of the interested estimator can be controlled. However, how to apply this interesting technique to our NGD algorithm seems not immediately straightforward. It should be an interesting topic for future study.

Next, note that Theorem  \ref{thm:gls}  is developed using standard asymptotic analysis techniques. The merit of this approach is that elegant theoretical results can be obtained in a more convenient way, with the help of a large sample size and infinite iterations. We next develop a parallel but non-asymptotic theoretical result with both finite sample size $n$ and number of iterations $t$. To this end, denote $\delta^0_{\max} = \max_{1 \leq m \leq M} \| \wh \theta^{(0,m)} - \thetaopt \|$ represents the initial distance. This leads to the following non-asymptotic results.

\begin{corollary}
\label{col:non-asymp}
(Non-Asymptotic Estimation Error)
Under the conditions of Theorem \ref{thm:gls}, there exist constant $c_3>0$, such that
\beqr
\label{eq:non-asymp}
\frac{\|\widehat{\theta}^{*(t)} - \thetaopt^*\|}{\sqrt{M} } \leq  \delta^0_{\max} ( 1 - \alpha \kappa_3)^t +    c_3 \big\{ \widehat{ \operatorname{SE}} (W) +   \alpha \big\} \big\{  \SEL +  \| \thetaopt - \theta_0  \| \big\}.
\eeqr
\end{corollary}
\noindent
The detailed proof of Corollary {\ref{col:non-asymp}} is given in Appendix B.4. By (\ref{eq:non-asymp}), we are able to decompose the estimation error of $\wh \theta^{*(t)}$ into three components. They are, respectively, the optimization error $\delta^0_{\max} ( 1 - \alpha \kappa_3)^t$, the  statistical error due to network structure, learning rate, and data distribution pattern $\big\{\SEW + \alpha \big\} \SEL,$ and the statistical error of the global estimator $\|\thetaopt - \theta_0 \|$. Note that, the term $\delta^0_{\max} ( 1 - \alpha \kappa_3)^t$  is the optimization error. It linearly converges to 0 as $t$ increases, which is consistent with the classical gradient descent methods \citep{nesterov1998introductory,boyd2004convex,karimi2016linear}.
Moreover, if the sample size is sufficiently large to support the asymptotic statistical inference, the statistical error of the global estimator $\|\thetaopt - \theta_0 \|$ should be of  $O_p(1/\sqrt{N})$ order. When both the sample size $n$ and number of iterations $t$ go to infinity, this result degenerates to that of Theorem 3.

\csubsection{Locally Over-Parameterized Problems}
\noindent

We study an interesting problem about the locally over-parameterized models in this subsection. By  local over-parameterization, we mean that the number of parameters (i.e., $p$) is larger than the locally averaged sample size $n$ but smaller than the whole sample size $N$. In contrast  to a locally over-parameterized problem, a globally over-parameterized problem refers to the situation with $N>p$. In this case, the stable solution cannot be uniquely determined. This makes the discussion about the resulting estimator's theoretical statistical efficiency  impossible. Consequently, we would like to focus on the locally over-parameterized problem in this subsection.

For illustration purpose, we consider here a linear regression model but under a locally over-parameterized framework (i.e., $n<p<N$). Following the same technique in Subsections 2.1 and 2.2, we find that whether the NGD algorithm numerically converges to the stable solution is fully determined by the contraction operator $\Delta^{*}(W \otimes I_p)$. Once again, whether $\lambda_{\max}\big\{\Delta^{*}(W \otimes I_p)\big\} < 1$ is the key condition. Nevertheless, this problem becomes considerably more challenging with $n>p$. In this case, we have $\lambda_{\max} \big ( W \otimes I_p \big) = 1$  and $\lambda_{\max} \big ( \Delta^{*}\big) = 1$. Consequently, there indeed exists the theoretical possibility that  $\lambda_{\max}\big\{\Delta^{*}(W \otimes I_p)\big\}= 1$; see Appendix C.1 for an example.  If this happens, then the numerical convergence of the NGD algorithm can no longer be assured. Thus, the key question here is: how likely this undesirable situation can happen in real practice. To address this interesting question, we study here two important network structures very carefully. They are, respectively, the central-client network and the circle network.

{\sc Case 1 (Central-Client Network)}. As our first example, we consider here the central-client network structure. Recall that $W  = [0,(M-1)^{-1} \1^\top_{{_{M-1}}};\1_{{_{M-1}}},\textbf{0}].$  We  can prove that for any eigenvalue $\lambda$ of $\Delta^{*}(W \otimes I_p)$, there exists $\boldsymbol{\nu}_1\in \mR^p,$ such that
$(M-1)^{-1}\Delta^{(1)}\sum_{k=2}^M\Delta^{(k)}\boldsymbol{\nu}_{1} = \lambda^2 \boldsymbol{\nu}_{1};$ see detailed proof in Appendix B.6.  As a consequence, it suffices to verify that $\lambda_{\max}\big(\Delta^{(1)}(M-1)^{-1}\sum_{k=2}^M\Delta^{(k)}\big) < 1.$ Then it could be proved that
\beqrs
\Delta^{(1)}(M-1)^{-1}\sum_{k=2}^M\Delta^{(k)} &=& \left( I_p - \alpha \wh \Sigma_{xx}^{(1)} \right)  \left\{ I_p - \alpha  (M-1)^{-1} \sum_{m=2}^M \wh \Sigma_{xx}^{(m)} \right\} \\
&=& I_p - \alpha\left\{ \frac{M}{M-1} \wh \Sigma_{xx} + \frac{M-2}{M-1} \wh \Sigma^{(1)}_{xx} \right\} + O(\alpha^2) .
\eeqrs
Note that $\wh\Sigma_{xx}$ is a positive definite matrix, and $\wh\Sigma_{xx}^{(1)}$ is a semi-positive definite matrix. Thus, when $0<\alpha < 2(M-1)/\big\{ M \lambda_{\max}(\wh\Sigma_{xx}) + (M-2) \lambda_{\max}(\wh\Sigma_{xx}^{(1)})  \big\},$
the largest  absolute eigenvalue of the leading term of $\Delta^{(1)}(M-1)^{-1}\sum_{k=2}^M\Delta^{(k)}$ (i.e., $I_p - \alpha (M-1)^{-1} \big\{ M \wh \Sigma_{xx} + (M-2) \wh \Sigma_{xx}^{(1)}\big\}$) is smaller than $1$. Therefore, as long as $\alpha$ is sufficiently small, such that the $O(\alpha^2)$ is ignorable, we would have $\lambda_{\max}\big\{\Delta^{*}(W \otimes I_p)\big\} < 1$.

{\sc Case 2  (Circle Network)}. Next, we consider  circle-type network in (\ref{eq:circle}) but with $D = 1$. Similarly, we can prove that
for any eigenvalue $\lambda$ of $\Delta^{*}(W \otimes I_p)$, there exists $\boldsymbol{\nu}_1\in \mR^p,$ such that
$\big(\prod_{m=1}^M \Delta^{(m)} \big) \boldsymbol{\nu}_{1}^{(p)}= \lambda^M \boldsymbol{\nu}_{1}^{(p)}$. In addition, $\big(\prod_{m=1}^M \Delta^{(m)} \big)$ could be expanded as follows
\beqrs
\prod_{m=1}^M \Delta^{(m)}  &=& \prod_{m=1}^M  \left( I_p - \alpha \wh \Sigma_{xx}^{(m)} \right) = I_p - \alpha  M \wh \Sigma_{xx} + O( \alpha^2)  .
\eeqrs
Consequently, $\lambda_{\max}\big(\prod_{m=1}^M \Delta^{(m)}\big) < 1$ as long as  $\alpha$ is sufficiently small.

Subsequently, we consider building a simulation study to check the numerical convergence of the NGD algorithm under more network structures (including the interesting counter-example we constructed).
The detailed setting and results are shown in Appendix C.1. From the results, it could be found that
in the locally over-parameterized regime, the proposed NGD algorithm could still converge numerically under most of the network structures.

\csection{NUMERICAL STUDIES}

\csubsection{The Basic Setups}
\noindent

We present here a number of numerical studies to demonstrate the finite sample performances of the NGD method based on both simulation studies and a real data example. For simulation studies, several simulation models are used to generate the whole sample data. Once the sample data are generated, they are then distributed to different clients. We consider two different distribution patterns. The first one is { \it homogeneous}. All the observations are distributed to different clients randomly. Thus, the sample distribution across different clients should be homogeneous. The second one is {\it heterogeneous}. We sort all the observations according to their response values. Then, we sequentially distribute them to different clients according to their sorted response values. Thus, the observations allocated to the same clients share similar response values, and then, distributions should be heterogeneous across different clients. As a result, the estimators produced by different clients separately can  hardly be consistent. Once all the simulated data are generated and distributed across different clients, the clients are then connected with each other by one particular type of network structure, as described in Subsection 2.4.

For the entire simulation study, we fix the whole sample size as $N = 10,000$ and the number of clients as $M = 200$. Thus, the sample size for each client is given by $n = N/M = 50.$ For each simulation study, the experiments are replicated for a total of $R = 500$ times for each parameter setup. Let $\widehat{\theta}^{*(t)}_{r}$ be the estimator obtained in the $r$th replicate ($1 \leq r \leq R$) on the $t$th iteration with the initial value given by $\widehat{\theta}^{*(0)}_{r} = \mathbf{0}$ for every $1 \leq r \leq R.$  We then compute its mean squared error (MSE) values averaged from the whole clients as $ \big\| \widehat{\theta}^{*(t)}_r - \theta^*_0\|^2 /M,$ where $\theta^*_0 = I^* \theta_0$ represents the stacked true parameters. This leads to a total of $R$ MSE values in log-scale. We then demonstrate the performance of the proposed algorithm for different models and network structures in terms of the $\log$(MSE) values.

\csubsection{Least Squares Loss Functions}

\noindent

\begin{figure}[h]
    \centering
	\setlength{\abovecaptionskip}{1pt}
	\subfigure{
		\includegraphics[width=0.95\columnwidth]{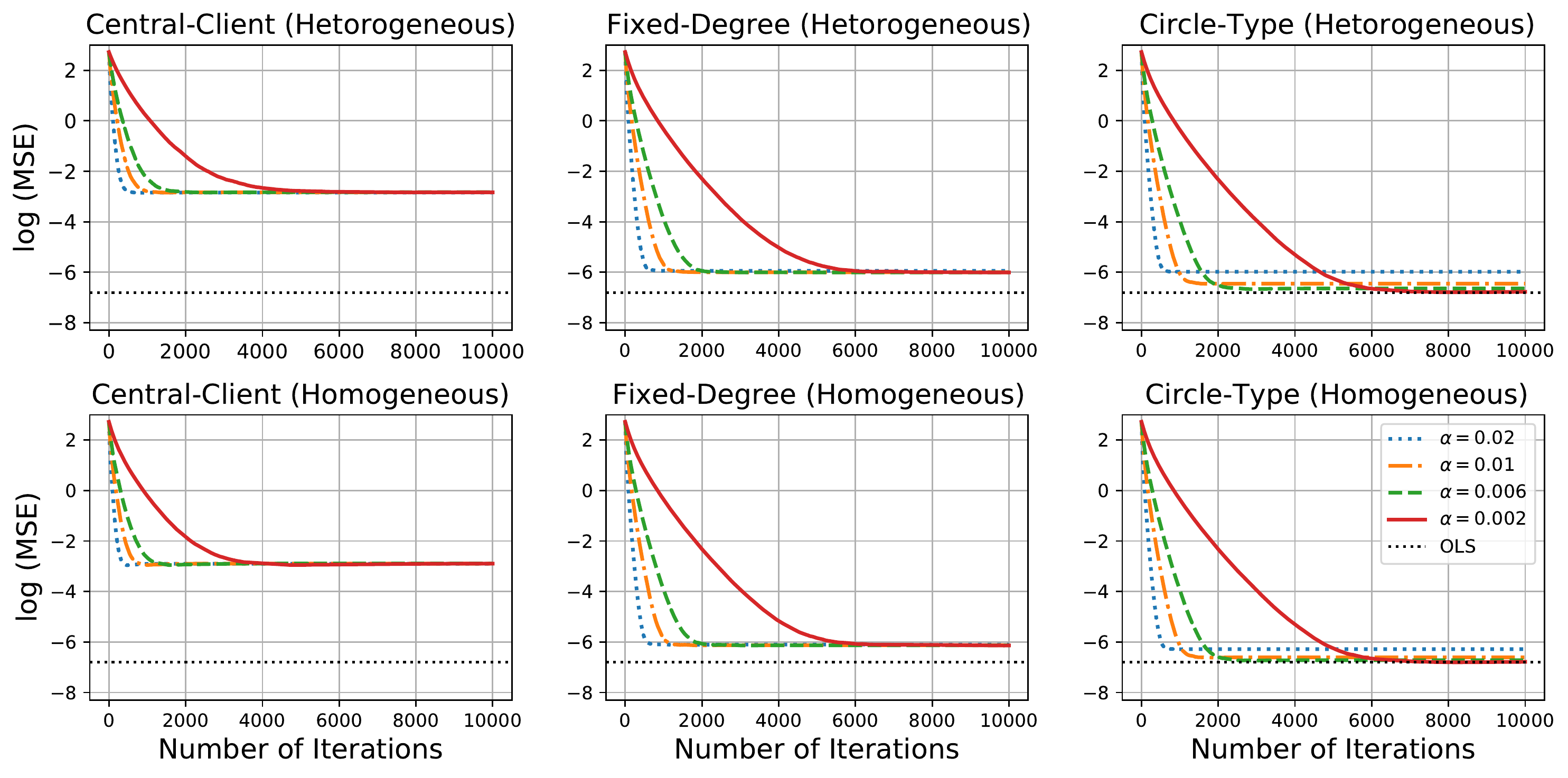} }
    \caption{The median log(MSE) values for different learning rates, network structures, and distribution patterns based on linear regression models. The black dotted line represents the global OLS estimators.}
    \label{fig:2}
\end{figure}

We start with a linear regression model. Following \cite{tibshirani1996regression}, we set $p = 8$ and $\theta_0 = (3,1.5,0,0,2,0,0,0)^\top $. Here, $X_i$ is generated from a multivariate normal distribution with mean 0 and $\operatorname{cov}(X_{ij_1},X_{ij_2}) = \rho^{|j_1 - j_2|}$ with $\rho = 0.5$ for $1 \leq j_1,j_2 \leq p.$ The residual term $\varepsilon_i$ is independently generated from the standard normal distribution. For comparison,  we fix the in-degrees in the circle-type and the fixed-degree network structures to be 1 and 2, respectively. Various learning rates (i.e. $\alpha = 0.005,0.01,$ $0.02,0.05$) are considered. Subsequently, the NGD estimators are computed. The median log(MSE) values are reported in Figure \ref{fig:2}.

By Figure \ref{fig:2}, we obtain the following interesting findings. First, we find that the circle-type network structure seems to be the best network structure, with the lowest median $\log$(MSE) values. This is not surprising, because it has the smallest $\SEW$ values as 0. Instead, the central-client network structure demonstrates the worst performance based on its largest $\SEW$ values. Unfortunately, this is one of the most popularly used network structures in practice. The fixed-degree network structure performs slightly worse than that of the circle-type network structure. Recall that the in-degree of the fixed-degree network structure is fixed to be 2 in this example. As demonstrated in the next subsection, its performance can be greatly improved by allowing the in-degree to be slightly larger. Second, comparing the upper and bottom panels,  we find that little difference is detected. This might be due to the fact $\alpha$ is set to be sufficiently small and $M$ is relatively large.  Third, for each network structure, we find that the smaller the learning rate $\alpha$ is, the slower the algorithm converges and the more efficient the NGD estimator is. All these results are in line with our theoretical findings in Theorem \ref{thm:numerical_converge} very well.

\csubsection{General  Loss Functions}
\noindent

In this subsection, we further demonstrate the finite sample performance of the NGD methods on different models and general loss functions. Compared with the previous subsection, the key diffidence is that the loss function is replaced by more general ones.
Specifically, we consider the negative two-times log-likelihood function as the general loss function with the following simulation examples.

{\sc Example 1. (Logistic Regression)} In this example, we consider a logistic regression, which is one of the most popularly used models for classification. Consider an example in \cite{barut2016conditional}. Set $p = 6$ and $\theta_0 = (1/2,1/2,1/2,1/2,1/2,-2.5/2)^\top.$ The covariate $X_i$ is generated from a multivariate normal distribution with $E(X_i) = 0,$ $\operatorname{var}(X_i) = 1,$ and $\operatorname{cov}(X_{ij_1},X_{ij_2}) = 0.5$ for $j_1 \neq j_2$. Given $X_i,$  the response $Y_i$ is then generated according to
$P(Y_i = 1 | X_i,\theta_0) = \{1 + \exp(- X_i^\top \theta_0)\}^{-1}.$ The learning rates are set to 0.02, 0.05, 0.1, and 0.2.

{\sc Example 2. (Poisson Regression)} This is an example revised from \cite{fan2001variable}. Specifically, the feature dimension is $p = 8$ and $\theta_0 = (1.2,0.6,0,0,0.8,$  $0,0,0)^\top.$ The first six components of $X_i$ are generated in the same way as in Subsection 3.1 but with $\rho = 0.2$. The last two components of $X_i$ are independently and identically distributed as a Bernoulli distribution with probability of success 0.5. All covariates are standardized with mean 0 and variance 1. Conditional on $X_i$, the response $Y_i$ is generated from a Poisson distribution with $ E(Y_i|X_i) = \exp(X_i^\top \theta_0).$ The learning rates are set to 2$\times 10^{-4}$, 3$\times 10^{-4}$, 5$\times 10^{-4}$, and 8$\times 10^{-4}$.

\begin{figure}[!h]
    \centering
	\subfigure{
		\includegraphics[width=0.95\columnwidth]{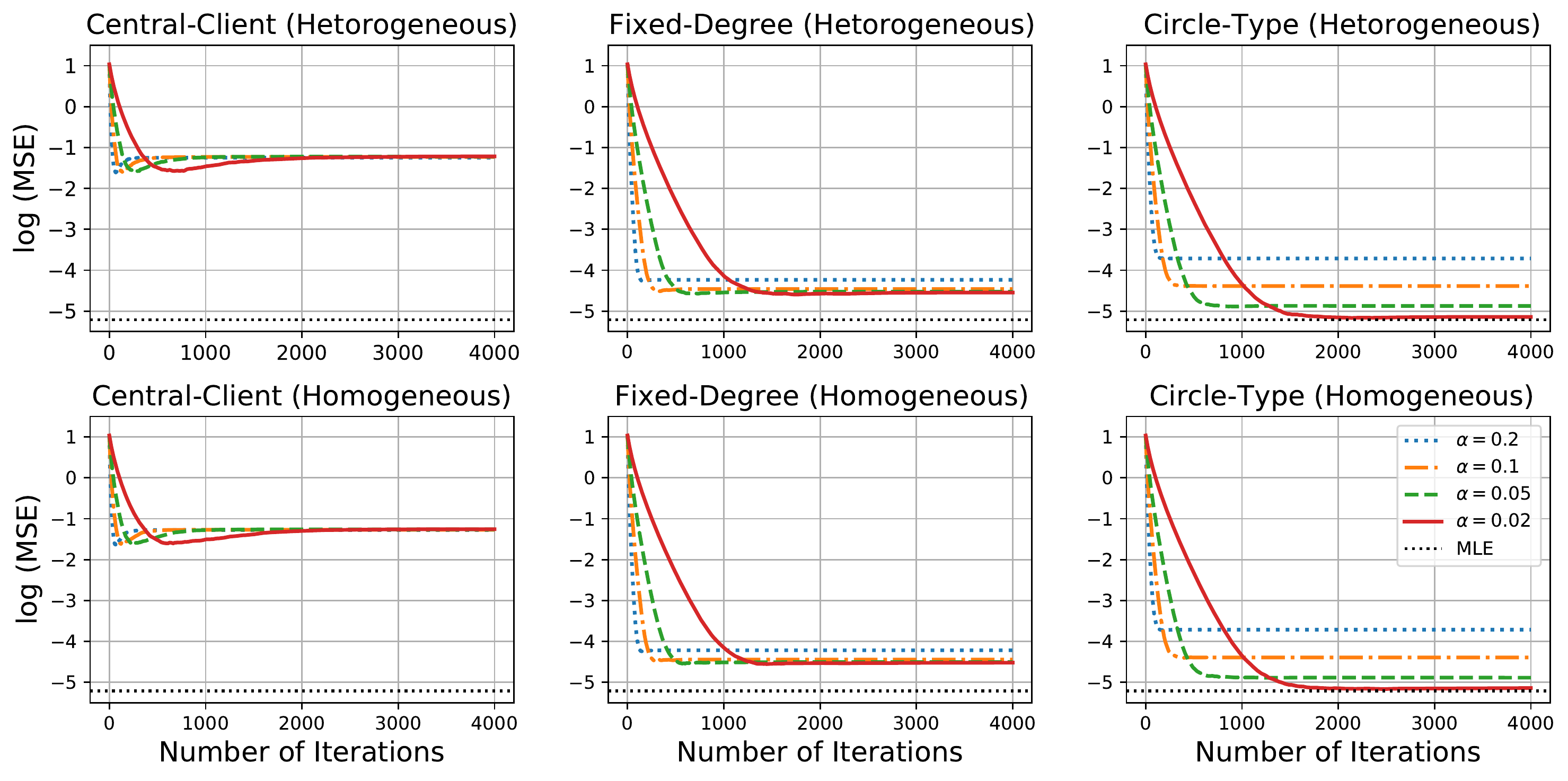} }
    \caption{The median log(MSE) values for different learning rates, network structures, and distribution patterns based on a logistic regression model. The black dotted line represents the global MLE estimators.}
    \label{fig:3}
\end{figure}

\begin{figure}[!h]
    \centering
	\subfigure{
		\includegraphics[width=0.95\columnwidth]{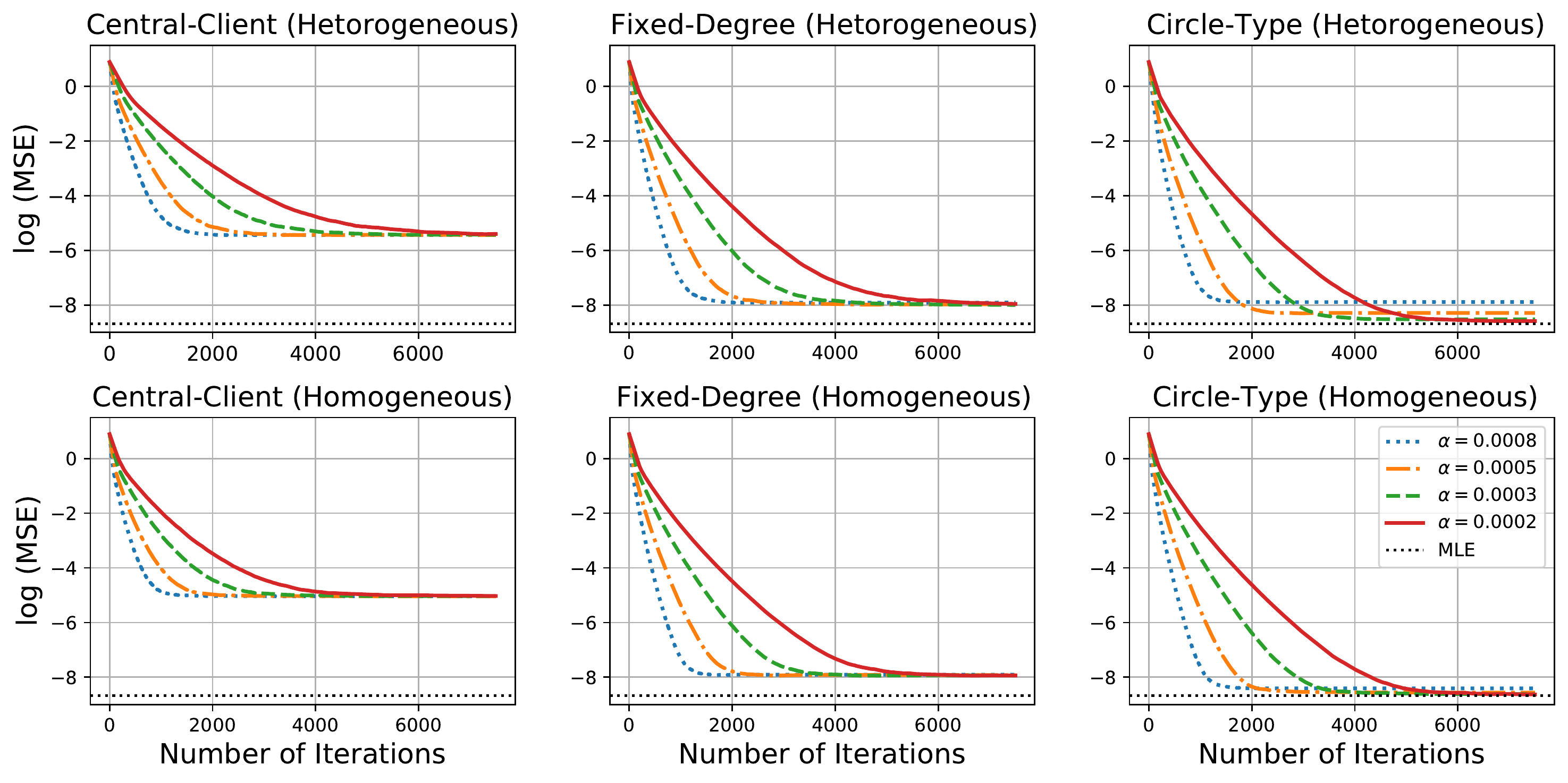} }
    \caption{The median log(MSE) values for different learning rates, network structures, and distribution patterns based on a Poisson regression model. The black dotted line represents the global MLE estimators.}
    \label{fig:4}
\end{figure}

The detailed results of the logistic regression are given in Figure \ref{fig:3},  while those of the Poisson regression are summarized in Figure \ref{fig:4}. All the results are qualitatively similar to those in Figure \ref{fig:2}. Specifically, we find that larger $\alpha$ leads to faster numerical convergence but worse statistical efficiency. Furthermore, the network structure plays a significant role in determining the statistical efficiency of the NGD estimators. The circle-type network structure is the best, and the central-client network structure is the worst. In addition, data heterogeneity has little influence on the statistical efficiency of the NGD estimators.

\csubsection{Nodal Degree Effect for Fixed Degree Networks}
\noindent

\begin{figure}[h]
    \centering
    \subfigure[Linear Regression]{\includegraphics[width=1.9in]{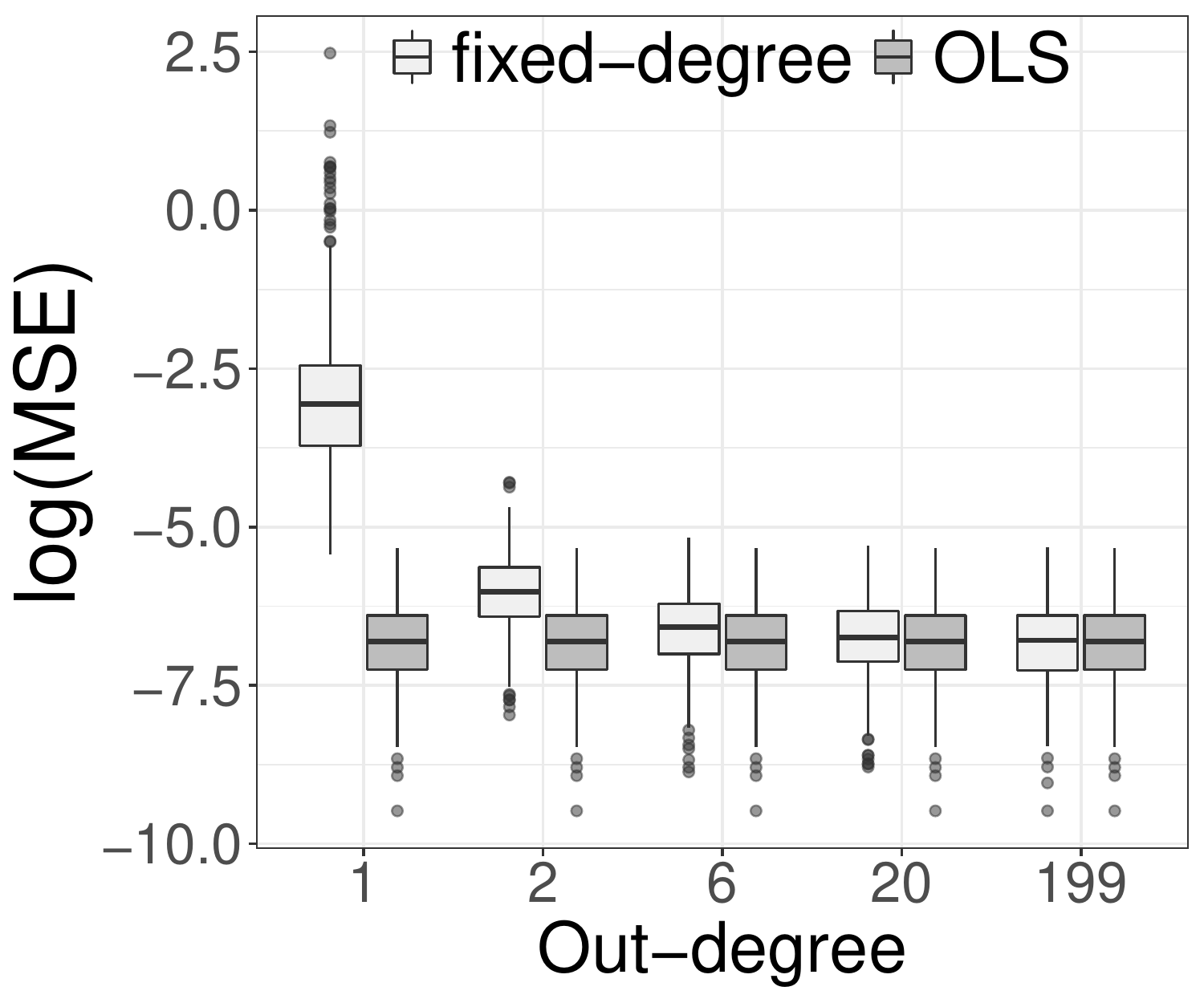} }
    \subfigure[Logistic Regression]{
        \includegraphics[width=1.9in]{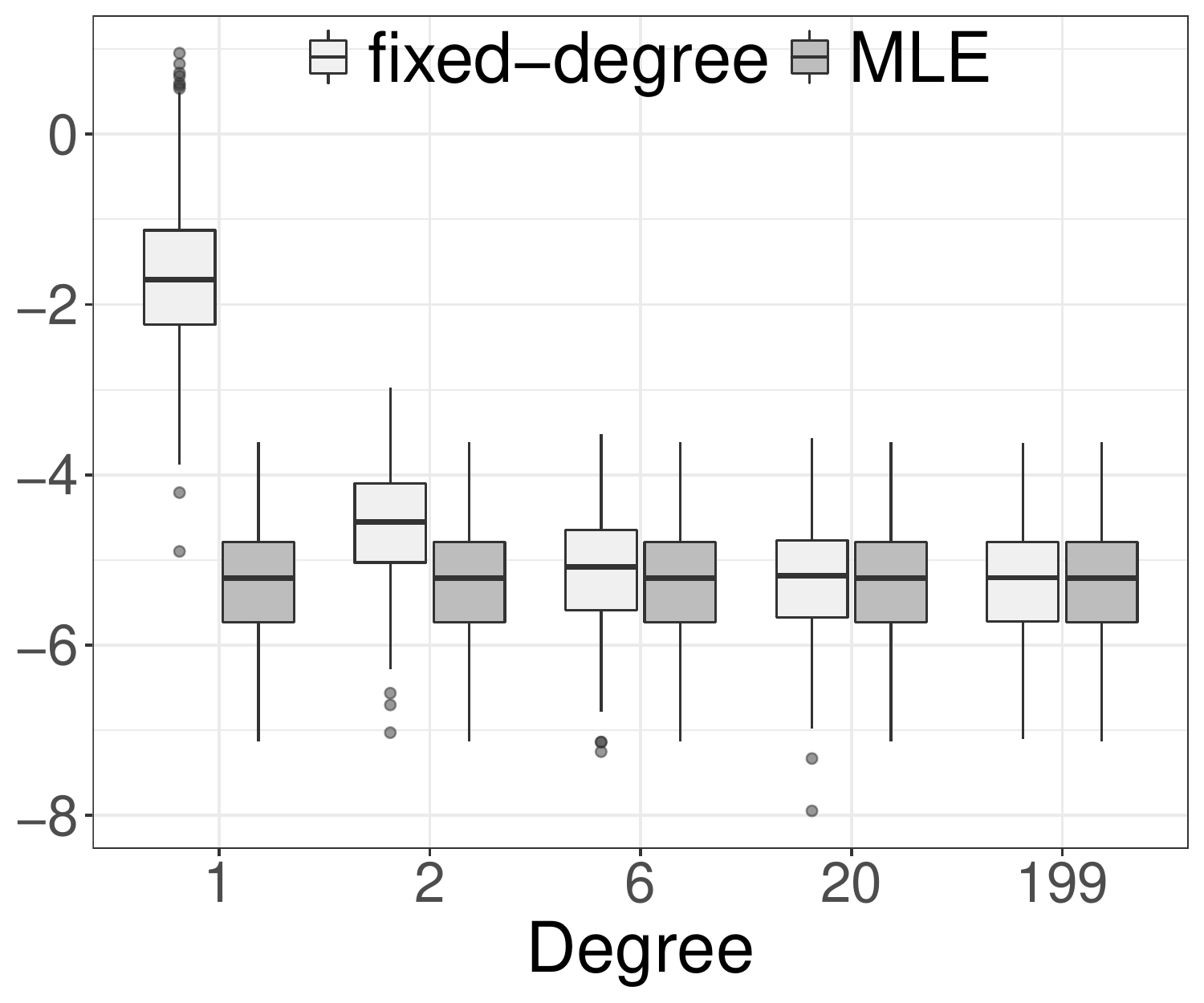}}
    \subfigure[Poisson Regression]{
        \includegraphics[width=1.9in]{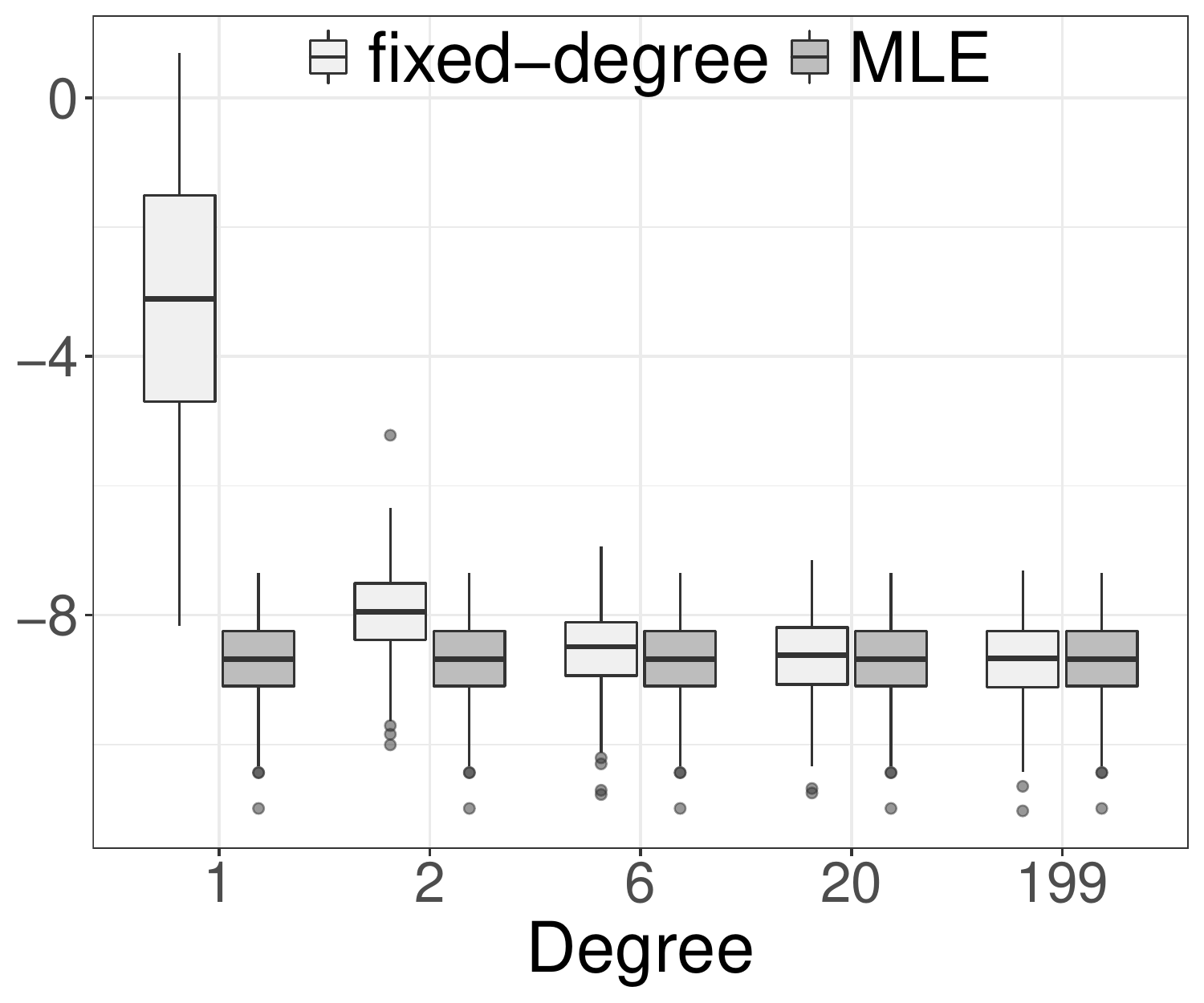}}
    \caption{The log(MSE) values for the fixed degree networks. The dark grey box represents the global estimator (OLS or MLE). From the left to the right panel, the learning rate is fixed to be $\alpha = 2\times 10^{-3}, 2\times 10^{-2} $, and $ 2 \times 10^{-4},$. }
    \label{fig:5}
\end{figure}

As mentioned in Subsection 2.3, the network structure plays an important role in determining the NGD estimators' statistical efficiency. Among various possible network structures, the fixed-degree type network is easy to implement in practice. Thus, we study its finite sample performance in this subsection. In particular, we focus on the effect of nodal degrees. Accordingly, the simulation example with the fixed-degree network structure in Subsection 2.4 is replicated for both OLS problem and general loss functions but with different nodal degrees. To ensure that the NGD algorithm converges to the global estimator, the learning rate considered for these three models is set to be sufficiently small as $\alpha = 2\times 10^{-3}$ for the linear regression, $2\times 10^{-2} $ for the logistic regression, and $ 2 \times 10^{-4}$ for the Poisson regression. The log(MSE) values are then box-plotted in Figure \ref{fig:5}.
We find that, as the in-degree increases, the statistical efficiency of the resulting estimator improves. In particular, the median log(MSE) decreases greatly when the in-degree increases from $1$ to $2$.
We also find that, when the in-degree is no less than $6$, the statistical efficiency of the resulting NGD estimator becomes very  comparable with that of the global estimator.

\csubsection{Deep Learning Models}
\noindent

In this subsection, we apply the proposed NGD method to more sophisticated deep learning models  for large-scale datasets. Specifically, we conduct the following two experiments.
The first experiment uses the MNIST dataset (\citealt{lecun1998gradient}, \url{ http://yann.lecun.com/exdb/mnist/}). It contains  70,000 photos of hand-written digits (0--9), and each class contains about 7,000 images.  60,000  of the dataset are for training and the rest are for validation.
For this dataset, we consider the LeNet model \citep{lecun1998gradient} with $p = 61,706$ parameters.  The second experiment studies the CIFAR10 dataset (\citealt{krizhevsky2009learning}, \url{http://www.cs.toronto.edu/~kriz/cifar.html}), which contains 60,000 colour images. The dataset forms 10 equal-size classes. Among the dataset, 50,000 are for training and 10,000 are for validation.
For this dataset, we consider the MobileNet model \citep{howard2017mobilenets}  with $p = 3,488,522$ parameters.

To train the model, we distribute the training data into $M$ clients with $M=40$ for MNIST and $M=25$ for CIFAR10. The data are sorted by response labels first and then distributed to different clients. Thus most of the clients contain only one class of response label, which is extremely heterogeneous across different clients.
 Three different network structures are studied, they are respectively: (1) the central-client network structure, (2) the circle type network structure with $D = 2$, and (3) the fixed degree network structure with $D = 6$.   To accelerate the numerical convergence, the constant-and-cut learning rate scheduling strategy is adopted \citep{wu2018understanding,lang2019using} for MNIST with initial learning rate $\alpha = 0.01$. Then, it drops to 0.005 and 0.001 after 1,000 and 4,000 iterations, respectively.
For CIFAR10, since MobileNet requires larger amount of GPU memory,
we utilize the mini-batch strategy \citep{cotter2011better,li2014efficient} with batch size 128 and initial learning rate 0.002.
Define one epoch to represent the process that the local gradient has been updated for every client based on all of its batches. The learning rate decreases to $0.001$ after 2,800 epochs. As to the initial value $\widehat\theta^{(0,m)}$, we adopt Xavier uniform initializer \citep{glorot2010understanding} for MNIST, and the pre-trained weights on ImageNet dataset for CIFAR10.

\begin{figure}[!h]
    \centering
	\subfigure[{\sc Lenent on MNIST}]{
		\includegraphics[width=0.9\columnwidth]{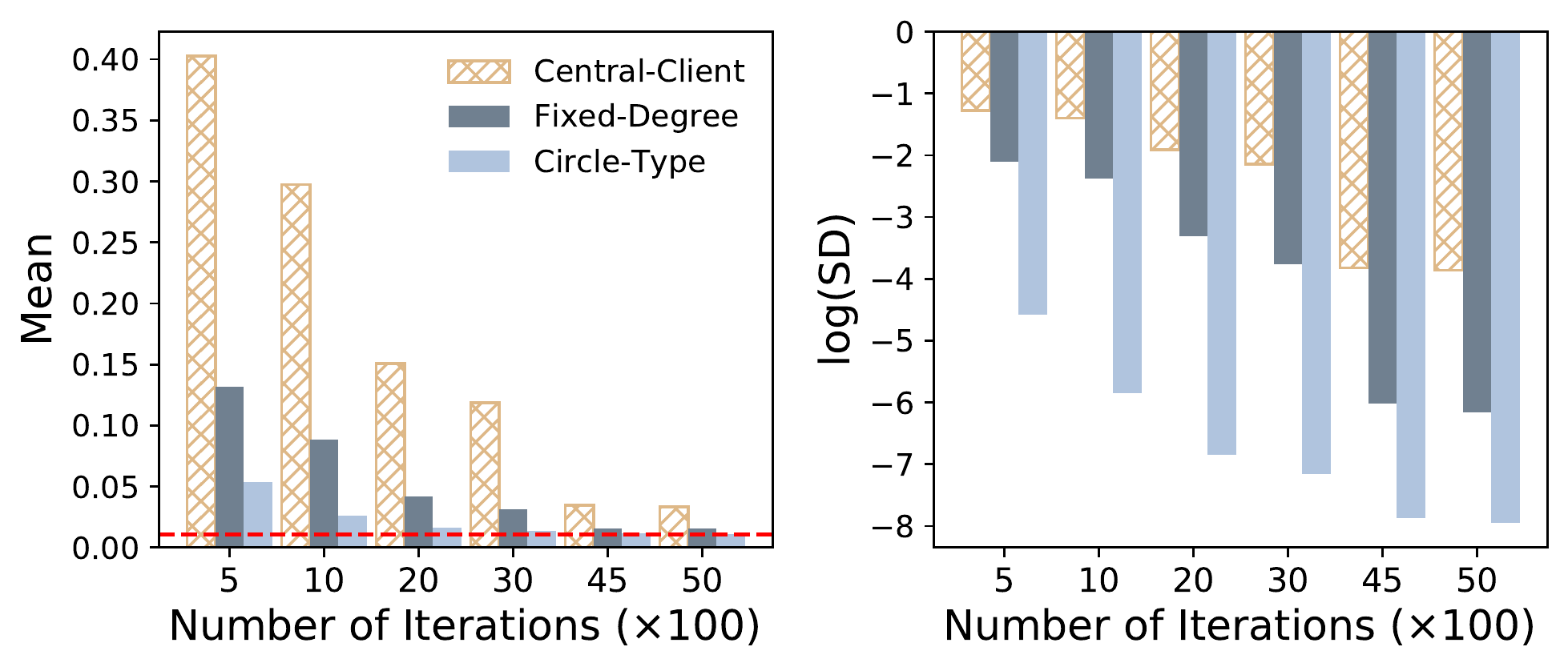} }
    \subfigure[{\sc MobileNet on CIFAR10}]{\includegraphics[width=0.9\columnwidth]{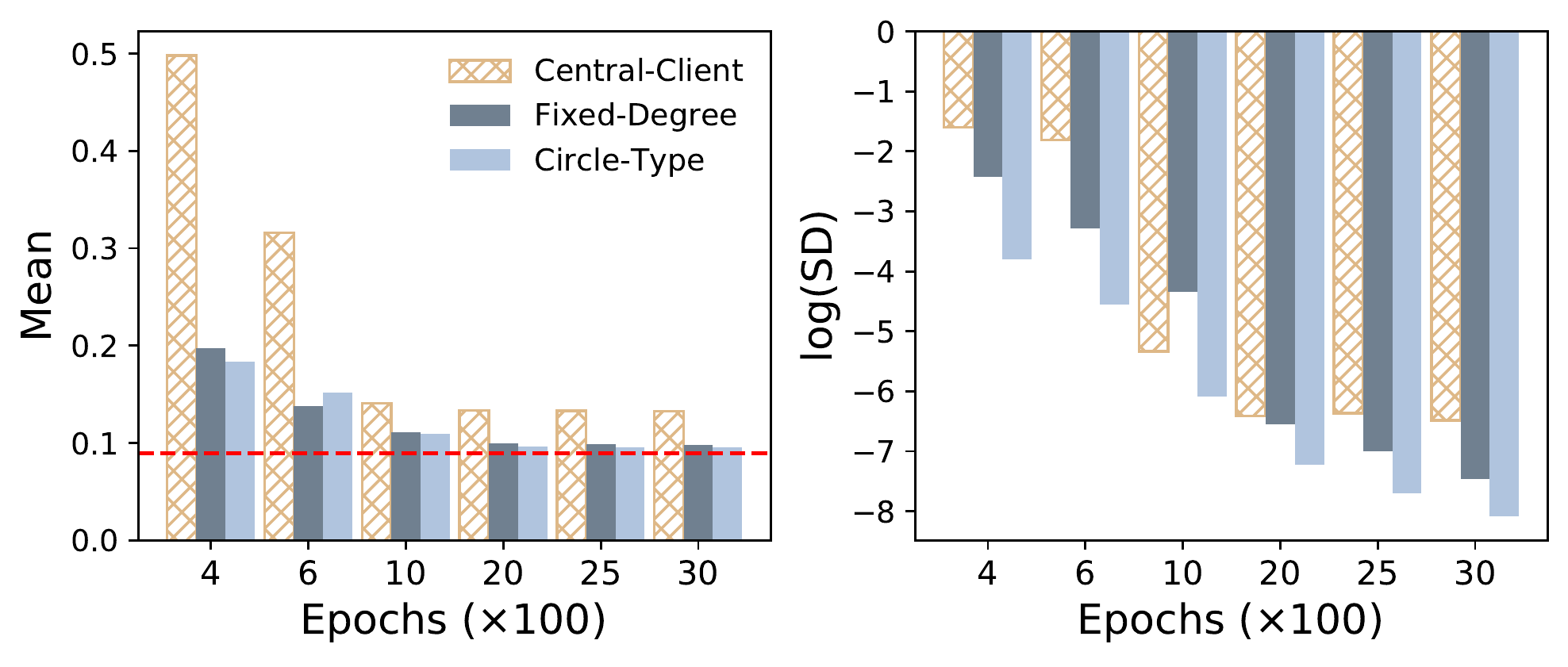}}
    \caption{ The mean and log(SD) for the Err values obtained by deep learning models. Different network structures are considered. The red dotted line in the left panel represents the optimal Err value calculated based on the whole dataset using one single computer.}
    \label{fig:barplot}
\end{figure}

The prediction performance of $\widehat\theta^{(t,m)}$ is then evaluated by the prediction error based on the validation set, which is denoted by $\operatorname{Err}^{(t,m)}$. Then, for any given network structure and a given $t$, we should obtain a total of $M$  $\operatorname{Err}$-values. Their means and log-transformed standard deviations are bar-plotted in Figure \ref{fig:barplot}. For comparison, we also report the {\it optimal Err value} calculated based on the whole dataset using one single computer. By Figure \ref{fig:barplot}, we obtain two interesting findings. First, from the left panel, we find that the mean Err values of the central-client type network structure are  the largest. In the contrast, those of both the circle-type and fixed-degree networks are  much smaller. By the time of convergence, the prediction errors  based on those two network structures could be as small as the optimal one.  While the circle-type network slightly outperforms the fixed-degree network. Second, from the right panel, it could be concluded the standard deviation of the Err values (in log-scale) for the central-client network is always the largest, while that of the circle-type network is the smallest.

\csection{CONCLUDING REMARKS}
\noindent

In this study, we develop a methodology for a fully decentralized federated learning algorithm. Both the theoretical and numerical properties of the algorithm are carefully studied. We find that the numerical convergence properties are mainly determined by the learning rate. However, the statistical properties of the resulting estimators are related to the learning rate, network structure and data distribution pattern. A sufficiently small learning rate and balanced network structure are required for better statistical efficiency, even if data are distributed heterogeneously. Extensive numerical studies are presented to demonstrate the finite sample performance.

Finally, we discuss interesting topics for future study. First, our study analyses a synchronous NGD algorithm, which ignores asynchronous problems, which should be the subject of further study. Second, our theorem suggests that a sufficiently small learning rate should be used for the best statistical efficiency. However, in practice, an unnecessarily small learning rate would lead to painfully slow numerical convergence. Then, an interesting research problem is how to practically schedule the learning rate to balance between statistical efficiency and numerical convergence speed.  In addition, our methodology focuses on the situation when the loss function is globally or locally strong convex. How to generalize our theoretical results to more general locally convex settings \citep{ho2020instability,ren2022towards} remains to be a challenging but also an interesting topic for future study.

\renewcommand \refname{\centerline{REFERENCES}}
 \bibliographystyle{asa}
\bibliography{ref}

\end{document}